\documentclass[runningheads]{llncs}
\usepackage{graphicx}
\usepackage{graphicx}
\usepackage{comment}
\usepackage{amsmath,amssymb} 
\usepackage{color}

\usepackage{ amsmath,epsfig}

\usepackage{booktabs}
\usepackage{amssymb}
\usepackage{amsmath}
\usepackage{graphicx}
\usepackage{bm}
\usepackage{color}
\usepackage{algorithm,algorithmic} 
\usepackage{epsfig}
\usepackage{caption} 
\usepackage{subfigure}
\usepackage{lipsum}
\usepackage{epstopdf}
\usepackage{array} 
\usepackage{bbding}
\usepackage[american]{babel} 
\usepackage{cite}
\usepackage{CJK}
\usepackage{footnote}
\usepackage[colorlinks,linkcolor=blue]{hyperref}

\begin{document}
\title{Novel View Synthesis on Unpaired Data by Conditional Deformable  Variational Auto-Encoder}
%
%

\author { 
Mingyu Yin \inst{1}
\quad Li Sun\inst{1,2}\thanks{Corresponding author, email: sunli@ee.ecnu.edu.cn.
Supported by the the Science and Technology Commission of Shanghai Municipality (No.19511120800).}
\quad Qingli Li\inst{1}
}


\authorrunning{F. Author et al.}
%
\institute{Shanghai Key Laboratory of Multidimensional Information Processing, 
\\
\and
Key Laboratory of Advanced Theory and Application in Statistics \& Data Science,\\
East China Normal University, 200241 Shanghai, China
} 
%
\maketitle              
\begin{abstract}
Novel view synthesis often needs the paired data from both the source and target views. 
This paper proposes a view translation model under cVAE-GAN framework without requiring the paired data. We design a conditional deformable module (CDM) which uses the view condition vectors as the filters to convolve the feature maps of the main branch in VAE. It generates several pairs of displacement maps to deform the features, like the 2D optical flows. The results
are fed into the deformed feature based normalization module (DFNM), which scales and offsets the main branch feature, given its deformed one as the input from the side branch.  Taking the advantage of
the CDM and DFNM, the encoder outputs a view-irrelevant posterior, while the decoder takes the code drawn from it to synthesize the reconstructed and the view-translated images. To further ensure the disentanglement between the views and other factors, we add adversarial training on the code. 
The results and ablation studies on MultiPIE and 3D chair datasets validate the effectiveness of the framework in cVAE and the designed module. \url{https://github.com/MingyuY/deformable-view-synthesis}
\end{abstract}
\section{Introduction}
Based on only a few sample images of a certain object with different poses, humans have the strong ability to infer and depict 2D images of the same object in arbitrary poses \cite{shepard1971mental}. This paper focuses on a similar task, known as the novel view synthesis, which aims to make computer render a novel target view image of an object given its current source view input. Obviously, this task requires the computer to understand the relationship between the 3D object and its pose. It has many potential applications in computer vision and graphic such as action recognition \cite{wang2014cross}, 3D object recognition \cite{savarese2008view}, modeling and editing \cite{massa2016deep} \emph{etc.}. 
\begin{figure}
\centering
\includegraphics[width=1.0\textwidth]{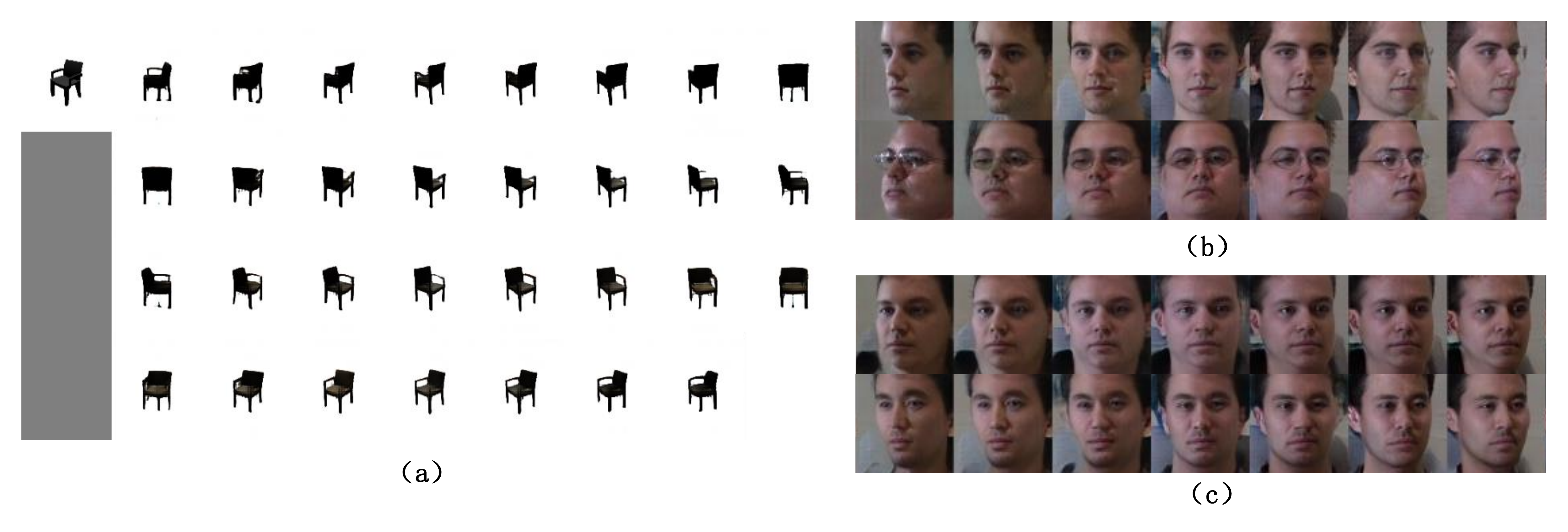}
\caption{
We use unpaired data to realize view synthesis. In (a), given the
first source view image, the chair rotates with a span of 360$^\circ$ . In (b), faces
are synthesized into existing predefined views in the dataset. In (c), we are able to interpolate the face into unseen views in the training data. Details are given in the result section \ref{subsec:abl} and \ref{subsec:rel}.}

\label{fig:main0}
\end{figure}
Traditional approaches \cite{avidan1997novel,kholgade20143d} for this task are mainly based on 3D projection geometry. They first construct the 3D shape model of the object from the cues in the image. Then the model is projected onto a 2D image plane of the target view. Actually, if 3D model can be perfectly built, object in arbitrary poses can be rendered precisely. However, building 3D object model from a single 2D image is an ill-posed problem. Therefore, it needs a large amount of close viewpoint images to capture the full object structure. Since structures of various objects are quite different, 3D geometry model for a particularly object may not generalize to other. Moreover, rendering a high quality image not only depends on the object model, but also other conditions such as the lighting and the background, but they need to be modeled independently.

Learning based approaches \cite{rematas2014image,zhou2016view} begin to show the advantages with the help of deep convolutional neural network (CNN). This type of methods directly learn the mapping network from the source view to the target without building the 3D model and knowing the camera pose. The mapping network is modeled by a huge number of parameters determined in the data-driven manner. Hence it is large enough to accommodate not just the geometry projection function, but the background and lighting conditions. 
Recently, employing the image generation technique like generative adversarial notwork (GAN) has drawn researchers' attention. 
\emph{E.g.}, novel synthesis can be modeled by a conditional GAN (cGAN) just like image-to-image translation \cite{isola2017image}. 

Disadvantages of such methods lie in two aspects. First, the model dose not consider the prior knowledge about the projection geometry, though, previous works \cite{tran2017disentangled} already achieve the promising results given both the pose and identity labels as conditions. The works in \cite{nguyen2019hologan,sun2018multi,xu2019view} improves this by either designing differentiable 3D to 2D projection unit \cite{nguyen2019hologan}, predicting the warping flow between two different views \cite{sun2018multi}, or using a specific pose matrix rather than one hot vector as the input conditions \cite{xu2019view}. 
Training such a view translation model often requires the paired data, with one being used as the source view and the other as the target. The paired data essentially provide the important constraining loss function for minimization. Nonetheless, the ground truth data from target view are not easy to obtain in real applications. Lately, with the recent synthesis technique \cite{zhu2017unpaired,bao2017cvae}, building a translation model by unpaired data becomes possible, which can greatly release the constraint of novel view synthesis. 

This paper proposes a novel view synthesis algorithm using the conditional deformable flow in cVAE-GAN framework, and it designs for training with the unpaired data, although it still achieves the better results if the target view image can be further exploited in the loss functions. The key idea is to perform the view translation by deforming the latent feature map with the optical flows,  computed from by the image feature and the view condition vectors together. 
We find that cVAE is able to disentangle the view-relevant and irrelevant factors, by mapping different source view images into posteriors, and making them close to a common prior.
It greatly increases the performance on the unpaired data. To further improve the synthesis results, we incorporate the adversarial training in the pixel and latent feature domain, and the reconstruction loss on the sampling code from the view-irrelevant posterior.

Specifically, we built the generator with a pair of connected encoder and decoder. 
The source and target view conditions are added into them by our proposed conditional deformable module (CDM), in which the one-hot view vector is first mapped into two latent codes, and then they are used as two filters to convolve the features, giving the displacements on $x$ and $y$ directions. Note that instead of one flows, we actually get $3\times3$ flows for each location like in \cite{dai2017deformable}. To achieve this, the features are divided into $9$ channel groups and the two filters convolve each group to output a pair of displacement maps. Each $3\times3$ results then deform the corresponding location in its $3\times3$ neighbourhood, naturally followed by an ordinary conv layer to refine feature maps after the deformation. Rather than directly giving the deformed features into the later layers, we also design a deformed feature based normalization module (DFNM), which learns the scale and offset 
given the deformed feature as its input. With the help of the CDM and DFNM, the encoder maps the source into a posterior, while the decoder transforms the code, sampled from either the posterior or the prior, back into a target view image. Besides the reconstructed and prior-sampled image in traditional cVAE-GAN, our model also synthesizes a view-translated image to guide the generator for the view synthesis task.

The contributions of this paper lie in following aspects. First, we build a model in cVAE-GAN for novel view synthesis based on the unpaired data. With the traditional and the extra added constraining loss, the model maps the source image into a latent code, which does not reflect the view conditions. The target view then complements the code in the decoder. Second, we propose two modules named the CDM and DFNM for view translation. They fits in our model to improve the synthesis results. Third, extensive experiments are performed on two datasets to validate the effectiveness of the proposed method.


\section{Related Works}
\textbf{Image generation by VAE and GAN.} GAN \cite{goodfellow2014generative} and Variational Auto-Encoder (VAE) \cite{kingma2013auto} are two powerful tools for generating high dimensional structured data. Both of them map the random code drawn from the prior into the image domain data. GAN introduces a discriminator $D$ to evaluate the results from the generator $G$. $D$ and $G$ are training in the adversarial manner, and finally $G$ is able to synthesize high quality images. However, GAN's training is unstable, and mode collapse often happens. Therefore, extra tricks are often added to limit the ability of $D$ \cite{gulrajani2017improved,heusel2017gans}. VAE has a pair of encoder and decoder. In VAE, the input image is first mapped into the latent probabilistic space by the encoder. The decoder takes the random code drawn from the posterior to reconstruct the input image. VAE can be easily trained by the reconstruction loss together with KL loss as its regularization. But it tends to give the blurry image. So it usually works with a discriminator to form a GAN \cite{larsen2015autoencoding}. Originally, both GAN and VAE perform unconditional generation. To better control the generated results, cGAN \cite{mirza2014conditional,isola2017image,miyato2018cgans} and cVAE \cite{sohn2015learning,bao2017cvae} are proposed. In these works, the conditional label is given to the network as the input. So it controls the generation results to fulfill the required condition. $D$ in cGAN not only evaluates the image quality, but also the condition conformity.  
GAN and VAE become popular tool in novel view synthesis. Particularly, the latent code is disentangled into different dimensions in the unsupervised way \cite{higgins2017beta,nguyen2019hologan}, with some of them naturally controlling the pose, which shows their great potential on view synthesis.

\textbf{Novel view synthesis.}
Novel view synthesis is a classical topic in both computer vision and graphics. Traditional approaches are built by the 3D projection geometry \cite{avidan1997novel,savarese2008view,kholgade20143d,zhang2015meshstereo,rematas2016novel}. These approaches estimate the 3D representation of the object, including the depth and camera pose \cite{avidan1997novel}, 3D meshes \cite{zhang2015meshstereo} and 3D model parameters \cite{savarese2008view,kholgade20143d,rematas2016novel}. Learning based method becomes increasingly popular with the help of CNN. Since all types of 3D representations can now be estimated by CNN, it is the main building blocks of the view synthesis algorithm. Dosovitskiy \emph{et al.} \cite{dosovitskiy2015learning} learn a CNN which takes the low dimensional code including the shape and camera pose as the input, and maps it into a high dimensional image. Zhou \emph{et al.} \cite{zhou2016view} employ a CNN to predict the appearance flow to warp source view pixels directly. However, without the adversarial training, these works tend to give low quality images. 

Since GAN and VAE is able to generate high quality images, GAN-based method becomes dominant recently \cite{park2017transformation,tran2017disentangled,sun2018multi,tian2018cr,xu2019view}. Park \emph{et al.} \cite{park2017transformation} predict the flow and the occlusion map to warp pixels first, and then the deformed image is given to the following network for refinement. The work \cite{sun2018multi} fully exploits a sequence of source images by giving them to an RNN-based network, which predicts a series of warping flows from sources to the current target view. In DR-GAN \cite{tran2017disentangled}, a connected encoder-decoder based generator is proposed. The encoder transforms the image into a latent code. Together with the target view condition, the code is applied by the decoder to synthesize the image. The discriminator in DR-GAN takes advantage of the ID labels to ensure the view translation not to change the source ID. CR-GAN \cite{tian2018cr} extends the encoder-decoder based structure by adding an extra path beginning from the decoder, which gives an extra reconstruction constraint in the image domain. VI-GAN \cite{xu2019view} employs the estimated camera pose matrix as the input condition for both source and target views, which replaces the one-hot condition vector. It also feeds back the view-translated image into the encoder, and requires its latent code to be close with the code from the source view, hence building the view-independent space. Note that in the above works, most of them \cite{park2017transformation,sun2018multi,tian2018cr,xu2019view} ask for the paired data to form the loss function. Although, DR-GAN do not have this constraint, it still requires the ID label for training the discriminator. Our work is totally based on the unpaired data and it dose not need any ID label during training.  
\section{Method}
\subsection{Overview framework}
\begin{figure}
\centering
\includegraphics[height=7.0cm]{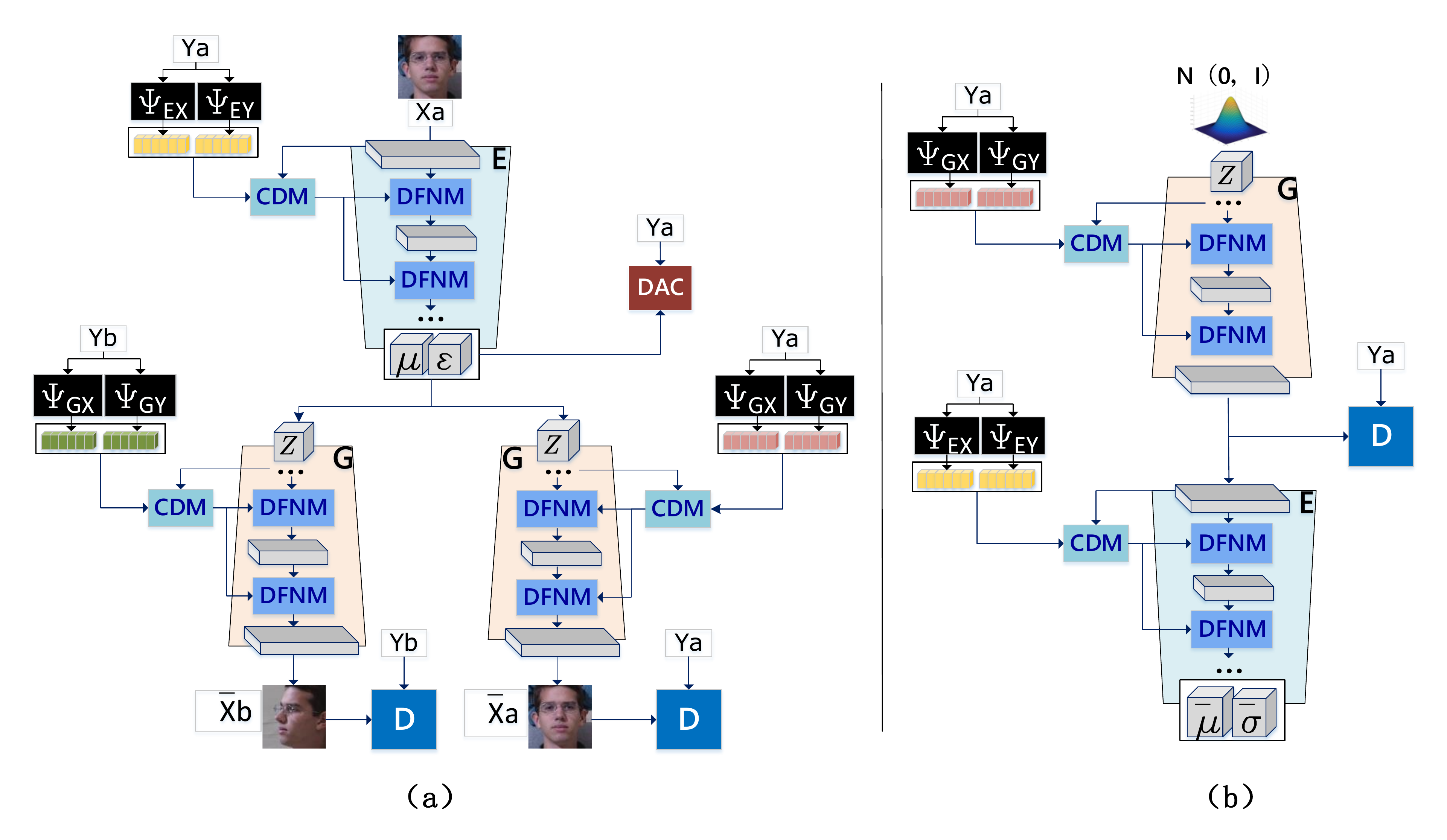}
\caption{Overview framework of the proposed network structure. (a) the source image $X_a$ with its label viewpoint $Y_a$ is translated into $\bar{X}_b$ in the target view $Y_b$. $\bar{X}_a$ is the reconstructed image with the same $Y_a$ given at both $E$ and $G$. (b) demonstrates that the code $Z\sim N(0,I)$ is synthesizing into a prior-sampled image, which is given back to $E$ to reconstruct the code $Z$.}
\label{fig:fig1}
\end{figure}

This paper regards the novel view synthesis as the condition translation task in cVAE-GAN. To achieve the view translation based on the unpaired data, we propose a conditional deformable module (CDM) and a deformed feature based normalization module (DFNM) in our designed network. To enhance the separation between the view-relevant and irrelevant factors, a disentanglement adversarial classifier (DAC) is also incorporated. As is shown in the Figure \ref{fig:fig1}, our network consists of three major components, an encoder $E$, a decoder $G$ and a discriminator $D$.  $\Psi_{EX}$, $\Psi_{EY}$ and $\Psi_{GX}$, $\Psi_{GY}$ are four different MLPs in $E$ and $G$, respectively. These MLPs maps the view label into conv filters, which are responsible for generating the optical flow. Given a source input image $X_a$ and its view label $Y_a$, the algorithm synthesizes a view-translated image $\bar{X}_b$ under the target view $Y_b$. Note that we do not have the ground truth $X_b$ to constrain the model during training.

In Figure \ref{fig:fig1},  
$E$ maps $X$ into a posterior $E(Z|X,Y)=N(\mu, \Sigma)$  
, from which a random code $Z\sim E(Z|X,Y)$ can be sampled. With $Z$ as its input, $G$ renders the fake images, and they are given to $D$ to evaluate the realness and view conformity. 
cVAE constrains $E(Z|X,Y)$ for all $X$ with the common prior $N(0, I)$ by reducing the KL divergence between them.  
 In cVAE, $E$ removes $Y_a$ from the source $X_a$, while $G$ adds $Y_b$ into the synthesized image. To fit the task of novel view synthesis, $G$ generates three kinds images: the reconstructed, prior-sampled images and the view-translated image. Note that, our model employs $Y$ as the input for $E$ and $G$. Instead of directly concatenation, we propose the modules CDM and DFNM, which make the whole network suitable for view translation.
Moreover, we follow the idea of BicycleGAN \cite{zhu2017toward} to reconstruct $Z$ from the prior-sampled image, and it ensures $G$ to take effective information from the code $Z$. 
\subsection{Conditional Deformable Module (CDM)}

We now give the details about the proposed CDM, applied in both $E$ and $G$. Our motivation is to change the source view $Y_a$ to the target $Y_b$ by warping $X_a$ with the optical flow. Therefore, the CDM actually learns to generate the 2D flows for the features. Note that the warping is particularly useful when $Y_a$ and $Y_b$ are close. However, if they are far from each other, the deformed feature needs to be refined and complemented by the later layers.


Here, we argue that the flows are mainly determined by $Y$, but they are also influenced by the content in $X$. Therefore, they should be computed from both of them. As the view label $Y$ has no spatial dimensions, 
$Y$ is first mapped into a latent code, and then the code convolves the feature to get the offsets. 
Specifically, two sets of MLPs, $\Psi_{EX}$, $\Psi_{EY}$ and $\Psi_{GX}$, $\Psi_{GY}$, first map $Y_a$ and $Y_b$ to the latent codes $W$ ($W_{EX}$, $W_{EY}$ in $E$ and $W_{GX}$, $W_{GY}$ in $G$). Here, we separate the filters for $x$ and $y$ directions, and for $E$ and $G$. Detailed discussions are given in the experiments. Then, $W$ are used as the filters to convolve on the feature maps, resulting several pairs of feature maps indicating the displacement $dx$ and $dy$ on $x$ and $y$ directions.
 
\begin{figure}
\centering
\includegraphics[width=1.0\textwidth]{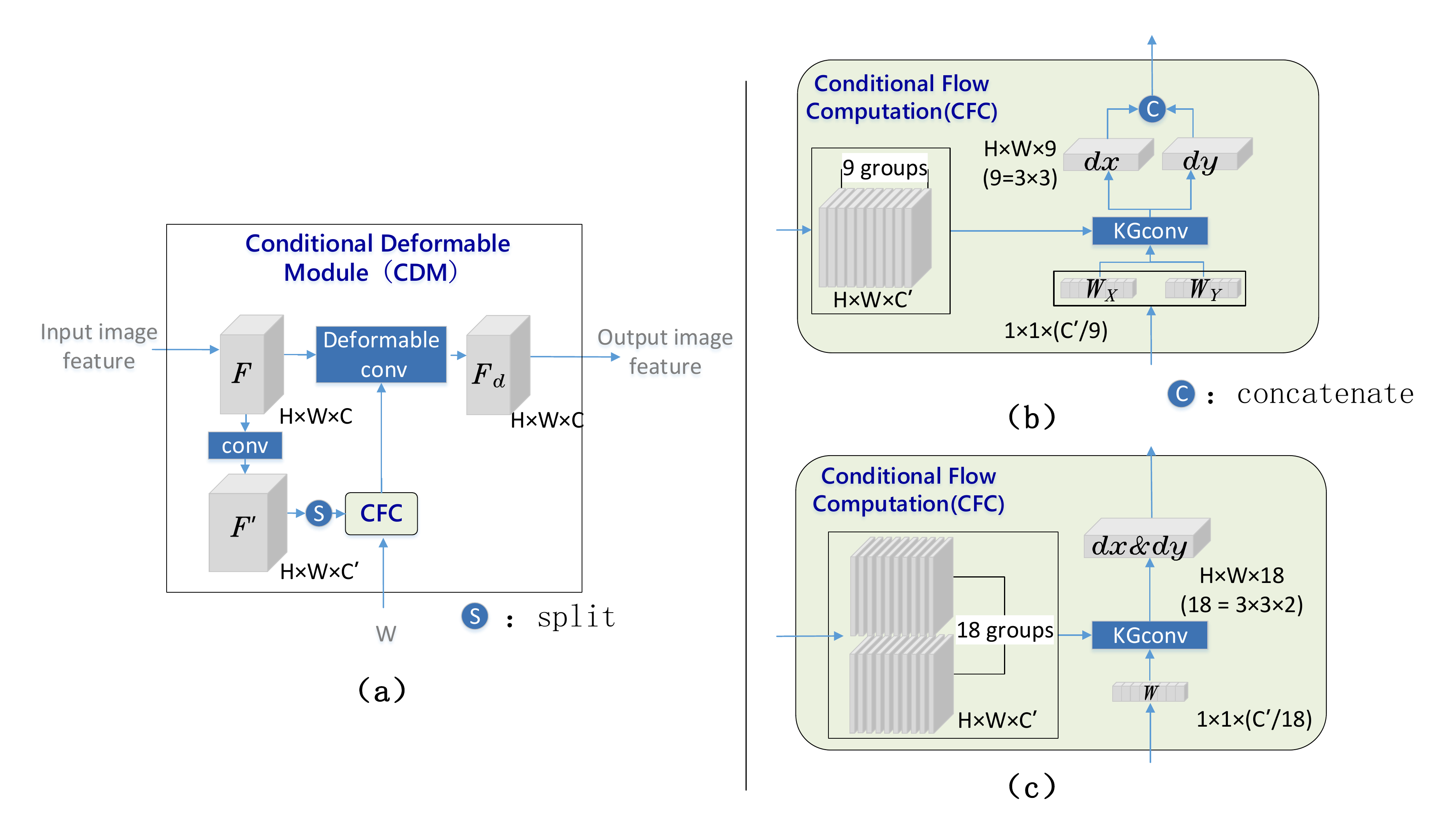}
\caption{The details for CDM. (a) Given the $F\in\mathbb{R}^{H\times W\times C}$ before the deformation, its output $F_d$ is the deformed feature with the same size as $F$. (b) CFC also has two separated input latent codes $W_X$ and $W_Y$, and they are used as the filters to convolve on a number (usually 9) of groups in $F'$. 
(c) Another design for CFC. Only one filter is provided, and it convolves on 18 groups.  }
\label{fig:fig2}
\end{figure}

Figure \ref{fig:fig2} shows the details about CDM. It mainly composed of the conditional flow computation (CFC) and the deformable conv module, as is shown in Figure \ref{fig:fig2} (a). Supposed the input $F^i\in \mathbb{R}^{H\times W\times C}$, of $i$th layer, CDM outputs the deformed $F_d^i$ of the same size. $W$ are also inputs, which are two latent vectors, computed from the view condition label $Y$ by MLP.
Particularly, $F^i$ is given to a conv layer with $C'$ filters to produce ${F'}\in \mathbb{R}^{H\times W\times C'}$. ${F'}$ is split into different groups along the channel, then given to the CFC. Figure \ref{fig:fig2} (b) and (c) are two options for CFC. In practice, we choose the design in Figure \ref{fig:fig2} (b), in which the layer of Kernel Given convolution ($KGconv$) uses $W_X, W_Y\in\mathbb{R}^{1 \times 1\times \frac{C'}{9}}$ as a pair of filters to convolve on each $\frac{C'}{9}$ intervals, leading to a pair of ${dx, dy}\in\mathbb{R}^{H\times W\times 9}$. Note that $dx, dy$ are composed of 9 groups of flows.  
Using 9 groups of flows is proposed by \cite{dai2017deformable} to introduce adaptive receptive fields in conv layer, 9 sets flows correspond to the offsets of a $3\times3$ conv kernels, and it finally gives the deformed feature ${F}_d^i$. We follow it but the flows are redundant and correlated to some extend, since they are the offsets of adjacent $3\times3$ elements. However, the 9 sets of flows could sometimes be different, depending on the data.

\subsection{Deformed Feature based Normalization Module (DFNM)}
The deformed feature maps ${F}_d^i$ need to be further processed by $(i+1)$th layers in $E$ and $G$. One intuitive way is to directly use ${F}_d^i$ as the input. However, recent advances in GAN and cGAN show the advantage of the conditional normalization like AdaIN \cite{huang2017arbitrary} and SPADE \cite{park2019semantic}. Different from BN or IN, such layers do not learn the scale $\gamma$ and offset $\beta$ as trainable model parameters. Instead, they are the features from the side branch. In other words, the conditional adaptive normalization module learns to scale and offset based on the conditional input. 

Inspired by SPADE, we propose a new conditional normalization way named DFNM, which uses ${F}_d^i$ as the conditional input from the side branch. DFNM performs the de-normalization, which means to determine the appropriate values on $\beta$ and $\gamma$. To be specific, it employs ${F}_d^i$ as its input, and specifies $\beta$ and $\gamma$ by two conv layers. Note that DFNM has distinct internal parameters for different layers, hence it progressively adjusts the features in the main branch based on its current input. In practice, we can have different choices on the dimensions of $\beta$ and $\gamma$. Here we simply follow the setting in SPADE, which outputs the unique $\gamma^i_{y,x,c}$ and $\beta^i_{y,x,c}$ at different 3D sites, where the subscripts are the indexes along the height, width and channel dimensions, respectively. Before the de-normalization, the features in the main branch should be normalized first by subtracting $\mu$ and dividing $\sigma$. 
Here we follow the way in BN to compute per-channel statistics 
$\mu^i_c$ and $\sigma^i_c$ from $h^i_{n,y,x,c}$ in the batch.


\subsection{Overall Optimization Objective}
The loss functions used in this paper mainly are three parts, namely, disentangling losses,  reconstruction losses and adversarial loss. 
\subsubsection{Disentangling loss} 
The disentangling loss constrains the encoder $E$, and prevents it from extracting the source view-relevant feature, so that the target view $Y_b$ can be easily added into the view-translated image. The KL constraint penalizes the posterior distribution $E (Z|X_a, Y_a)$ being far from the standard Gaussian $N(0,I)$, which to some extent makes the random code $Z\sim E(Z|X_a, Y_a)$ not carry the information related to $Y_a$.
KL loss $L_{KL}$, as is shown in Eq.\ref{eq:kl}, can be easily computed in closed form since both the prior and posterior are assumed as Gaussians.
\begin{equation}
\label{eq:kl}
L_{KL} = D_\text{KL}[E(Z|X_a, Y_a)||{N}({0},{ I})]
\end{equation}

However, this loss also constrains on view-irrelevant factors, so that this kind of information in $Z$ may lose because of the penalty from it. To cope with this issue, the paper proposes the DAC which mainly aims to reduce view-relevant factors in $Z$. With the help of DAC, the KL loss weight can be reduced so that the view-irrelevant factors remain in $Z$ to a greater extent. In practice, we implement the DAC as two FC-layers with the purpose of classifying the view based on $Z$. DAC is trained in the adversarial manner. Hence it has two training stages, $D$ and $G$ stages. 
In $D$ stage, the DAC is provided with the output $Z$ from $E$ and the correct source view label as well, while in $G$ stage, DAC is fixed and $E$ get trained with the adversarial loss from DAC. In this stage, we give an all-equal one-hot label to DAC 
with the same degree of confidence on each view. 
The cross entropy loss are defined as  Eq.\ref{eq:advE} and Eq.\ref{eq:advDAC}, respectively. 

\begin{equation}
\label{eq:advE}
L_{E}^{cls} = - \mathbb{E}_{Z\sim E(Z|X_a,Y_a)}\sum_c \frac{1}{C} \log DAC(c | Z)
\end{equation}
\begin{equation}
\label{eq:advDAC}
L_{DAC}^{cls} = - \mathbb{E}_{Z\sim E(Z|X_a, Y_a)}\sum_c \mathbb{I}(c=Y_a) \log DAC(c | Z)
\end{equation}
where $\mathbb{I}(c=Y_a)$ is the indicator function, and $DAC(c | Z)$ is softmax probability output by the disentanglement adversarial classifier. 
\subsubsection{Reconstruction losses} 
Reconstruction losses are important regularizations which also 
ensure that 
the view-relevant factors remain unchanged during view translation.
Without extra supervisions, cVAE wants the synthesized image $\hat{X}_a$ to be close to the input when $E$ and $G$ are provided the same view label $Y_a$. In addition,  
the constraints of the middle layer features of the classification network is also employed in our work. 
As shown in Eq.\ref{eq:L1 pixel} and Eq.\ref{eq:L1 gram}, $\phi^i$ indicates $i$th of a pre-trained VGG network, and $Gram$ means to compute the Gram matrix, which is a typical second order features.
\begin{equation}
\label{eq:L1 pixel}
\begin{aligned}
    L_{E, G}^{pixel} =& ||X_a -\bar{X_a}||_1,  \quad 
    L_{E, G}^{content} =& ||\phi^i({X_a}) -\phi^i({\bar{X_a}})||_1
\end{aligned}
\end{equation}
\begin{equation}
\label{eq:L1 gram}
\begin{aligned}
    L_{E, G}^{style} =& ||Gram(\phi^i({X_a})) -Gram(\phi^i({\bar{X_a}}))||_1 \\
\end{aligned}
\end{equation}
When $Z \sim {N} (0, I)$ for the prior-sampled image $G(Z, Y_a)$, we cannot constrain it directly in the image domain, so we extract the feature from the image $G(Z, Y_a)$ with $E$, and to reconstruct $Z$. 
So that the information in $Z$ is kept. The reconstruction loss expressed in Eq.\ref{eq:L1 z}
\begin{equation}
\label{eq:L1 z}
L^{rec_{z}}_{G} = \mathbb{E}_{Z\sim N(0, I)}||Z - E(G(Z, Y_a), Y_a)||_1
\end{equation}

\subsubsection{Adversarial loss}
In this paper, the projection discriminator \cite{miyato2018cgans} is adopted. 
Given the real image $X_a$, constraints are made for three types of fake images, reconstructed $G(E(X_a, Y_a), Y_a)$, view-translated $G(E(X_a, Y_a), Y_b)$ and prior-sampled image $G(Z, Y_a)$, as shown in Eq.\ref{eq:adv D} and  Eq.\ref{eq:adv EG}.
\begin{equation}
\label{eq:adv D}
\begin{aligned}
    L_{D}^{adv} =& \mathbb{E}_{{X}\sim p_{\rm{data}}}[\max(0,1-D(X,Y_a))]\\
    &+\mathbb{E}_{Z\sim E(Z|X_a, Y_a)}[\max(0,1+D(G(Z, Y_a)),Y_a)]\\
 &+\mathbb{E}_{Z\sim E(Z|X_a, Y_a)}[\max(0,1+D(G(Z, Y_b)), Y_b)]\\&+\mathbb{E}_{Z\sim N(0, I)}[\max(0,1+D(G(Z, Y_a)), Y_a)]
\end{aligned}
\end{equation}
\begin{equation}
\label{eq:adv EG}
\begin{aligned}
    L_{E, G}^{adv} =&\mathbb{E}_{Z\sim E(Z|X_a, Y_a)}[\max(0,1-D(G(Z, Y_a)),Y_a)]\\
 &+\mathbb{E}_{Z\sim E(Z|X_a, Y_a)}[\max(0,1-D(G(Z,  Y_b)), Y_b)]\\
&+\mathbb{E}_{Z\sim N(0, I)}[\max(0,1-D(G(z,  Y_a)), Y_a)]
\end{aligned}
\end{equation}
The total loss for $E$, $G$, $D$ and DAC can be written as following.
\begin{equation}
\label{eq:E all}
\begin{aligned}
   L_{E, G} = L_{KL} + L_{E, G}^{adv} + \alpha_1L_{E, G}^{style} +  \alpha_2L_{E, G}^{content} + \alpha_3L_{E, G}^{pixel} + L_{E}^{cls}   + L^{rec_{z}}_{G}
\end{aligned}
\end{equation}
\begin{equation}
\label{eq:E all}
\begin{aligned}
   L_{D} =  L_{D}^{adv},\quad
   L_{DAC} =  L_{DAC}^{cls} 
\end{aligned}
\end{equation}
We set the loss weight $\alpha_1= 0.001$,  $\alpha_2= 10$, $\alpha_3= 100$ for all experiments.
\section{Experiments}\label{sec:exp}

\subsection{Dataset and implementation details}
\textbf{Dataset.} We validate the proposed method on the 3D chair \cite{aubry2014seeing} and the MultiPIE face datasets \cite{gross2010multi}. The 3D chair contains $86,304$ images with a span of $360^\circ$ at azimuth and $30^\circ$ at pitch, respectively, covering a total of 62 angles. There are 1,392 different types of chairs. The multiPIE contains about 130,000 images, with a total span of $180^\circ$ and a spacing of $15^\circ$ in azimuth dimension. A total of 13 angles are used for training and testing. Meanwhile, it also contains images of 250 identities under different lights. For all the datasets, 80\% are used for model training and the rest 20\% for testing.

\begin{figure}
\centering
\includegraphics[height=8.0cm]{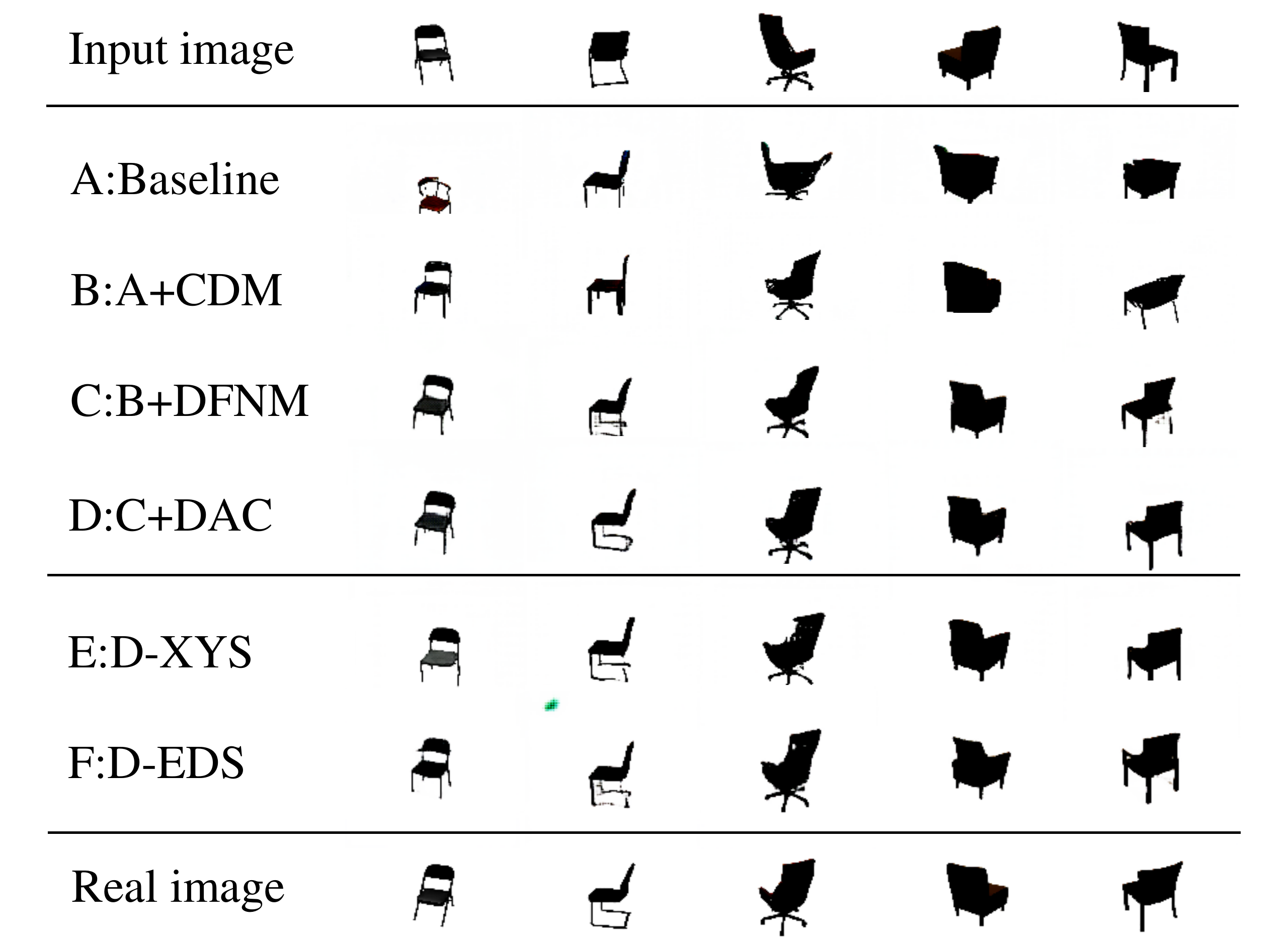}
\caption{Ablation study on 3D chair dataset.}
\label{fig:ablation}
\end{figure}

\textbf{Implementation details.}
In $E$ and $G$, all layers adopt instance normalization, except those replaced by DFNM. The spectral norm \cite{miyato2018spectral} is applied to all layers in $D$. All learning rates are set to 0.0002. We use the ADAM \cite{kingma2014adam} and set $\beta_1$ = 0, $\beta_2$ = 0.9. 
Details are given in the supplementary materials.

\subsection{Results and ablation studies on 3D chair and MultiPIE}\label{subsec:abl}
Extensive ablation studies is conducted to verify the effectiveness of each module. We have 6 different settings for it. View-translated images in different settings are presented in the corresponding rows in Figure \ref{fig:ablation} and the quantitative metrics are given in Table \ref{Table1}.
\begin{figure}
\centering
\includegraphics[width=1.0\textwidth]{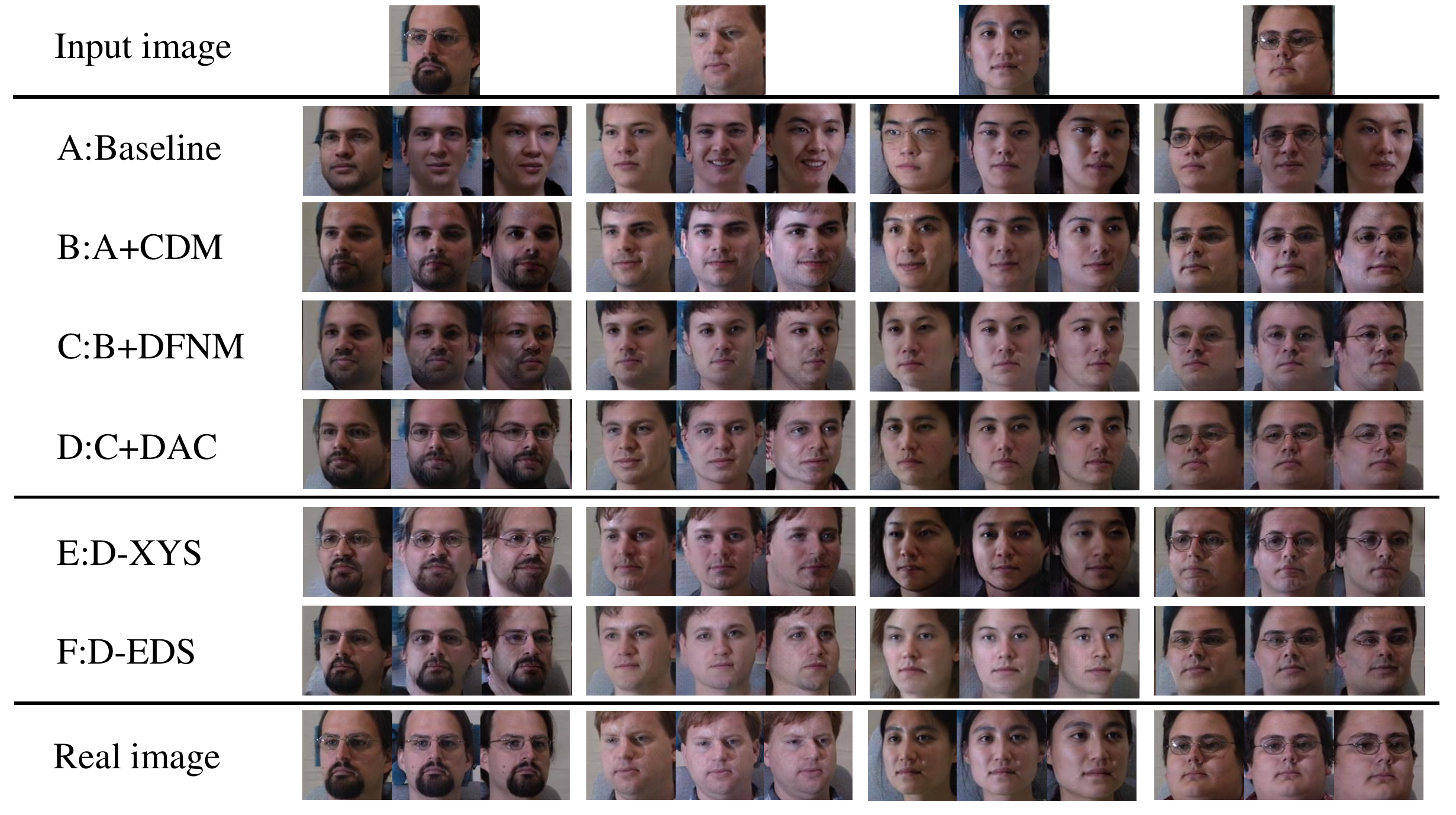}
\caption{Ablation study on multiPIE dataset.}
\label{fig:ablation multiPIE}
\end{figure}

\textbf{Baseline.} 
To verify the effectiveness of our proposed method, we use a general framework cVAE-GAN \cite{bao2017cvae} as the baseline. 
To make the comparison fair, we introduce the view-translated image in it, and use all the loss functions that is presented. The result is indicated as "A: baseline" in Table \ref{Table1} and Figure \ref{fig:ablation}, \ref{fig:ablation multiPIE}. 

\textbf{Validity of CDM.} 
To validate CDM, setting B is modified based on A. The only difference is we introduce the label through CDM, thus the setting is indicated by "B: A+CDM" in Table \ref{Table1} and Figure \ref{fig:ablation}, \ref{fig:ablation multiPIE}. 
Comparing the results between A and B in Figure \ref{fig:ablation}, we find that both can translate images to the given view. But when the difference between the target and input view is large, it is difficult for A to maintain the attributes and local details of the source image. While the CDM in B 
has the advantage of 
maintaining the representative details. 
In both the visual fidelity and similarity, B has a greater improvement on A. 

\textbf{Validity of DFNM.} 
We validate the DFNM in setting C based on B. The only difference between B and C is that we apply DFNM in C, while the deformed features are directly given to the later layers in the main branch in B. This setting is written as "C: B+DFNM" in Table \ref{Table1} and Figure \ref{fig:ablation}, \ref{fig:ablation multiPIE}. 
As is shown in Figure \ref{fig:ablation}, for some of the complex chair types, the synthesized image keep the chair style, indicating that DFNM helps catching the detail features in the source image. The quantitative results in Table \ref{Table1} indicate that DFNM refines the results compared with the setting B.

\textbf{Validity of DAC. } To demonstrate the effectiveness of DAC loss, we experiment in setting D based on C. In setting D, DAC is employed to provide the loss for encoder by Eq.\ref{eq:advE} . 
By introducing DAC, it enables $G$ to get more view-irrelevant information. In Figure \ref{fig:ablation}, \ref{fig:ablation multiPIE}, we can clearly see that although setting C basically maintain details, DAC in setting D gives a clearer representation. The results in Table \ref{Table1} give further proof that all metrics are improved on 3D chair, and L1 error and FID have only negligible decreasing on MultiPIE.



\textbf{Necessity of separating MLPs for $x$ and $y$ directions.} 
We are also interested in the way that CFC is implemented in CDM. There are at least two options for the filters $W$ from MLPs. One possible way is to employ the same $W$ to generate both $dx$ and $dy$, as is shown in Figure \ref{fig:fig2}(c). The other way is illustrated in the conditional flow computation sub-module in Figure \ref{fig:fig2}(b). The results of the first option are specified as "E: D-XYS", as is shown in Figure \ref{fig:ablation}, \ref{fig:ablation multiPIE} and Table \ref{Table1}.
We can see that 
the image is defective. The declines in quantitative metrics further illustrate the necessity of our design in CDM. 

\textbf{Necessity of separating the MLPs in $E$ and $G$.} $E$ and $G$ both use CDM to warp the features. But considering the different purposes of $E$ and $G$, the input conditional filters are different, coming from $\Psi_{EX}$, $\Psi_{EY}$, and $\Psi_{GX}$, $\Psi_{GY}$, as is shown in Figure \ref{fig:fig1}. 
We are wondering whether separating the MLPs in $E$ and $G$ is necessary, hence we implement a network in which $\Psi_X$, $\Psi_Y$ are sharing in $E$ and $G$. The results are presented as "F: D-EDS", 
which are worse than D, as is shown in Figure \ref{fig:ablation}, \ref{fig:ablation multiPIE} and Table \ref{Table1}. It shows the necessity of separating MLPs. 
\begin{table}[ht] 
\begin{center}
\setlength{\tabcolsep}{1mm}{
\begin{tabular}{ l l l c c c c c c c}
\toprule
& method && \multicolumn{3}{c}{MultiPIE}&&\multicolumn{3}{c }{3D chair}  \\
\cline{4-6}\cline{8-10}
&&&L1  &SSIM &FID&  &L1  &SSIM &FID  \\
\midrule
A:& Baseline &     	  & $31.37$  & $0.49$ &$44.84$&    & $8.39$   & $0.86$ &$104.78$  \\				
B:& CDM &      	  	  & $23.43$  & $0.55$   &$26.79$&   & $7.88$  & $0.87$ &$88.23$  \\				 	
C:& B + DFNM &      &\boldmath{$21.53$}  & $0.56$ &\boldmath{$23.59$}&   &$6.68$    & $0.88$ &$93.11$ \\				
D:& C + DAC&     &  $21.90$  & \boldmath{$0.57$} &$23.95$& & \boldmath{$6.37$} &\boldmath{ $0.89$} &\boldmath{$86.34$}   	\\			
\midrule
E:& D - XYS  &     	  & $24.48$  & $0.54$ &$31.02$&    & $7.18$   & $0.88$ &$90.31$  \\				
F:& D - EDS &      	  	  & $23.59$  & $0.54$   &$28.40$&   & $6.94$  & $0.88$ &$89.56$  \\			
\bottomrule
\end{tabular}
}
\end{center}
\caption{Quantitative ablation study on the MultiPIE and the 3D chair dataset. The pixel-wise mean L1 error and the structural similarity index measure (SSIM) \cite{wang2004image} are computed between the view-translated images and the ground truths. Besides, the FID is also reported.}
\label{Table1}
\end{table}

\subsection{Results and analysis on MultiPIE. }\label{subsec:rel}
\begin{figure}[ht]
\centering
\includegraphics[height=7.5cm]{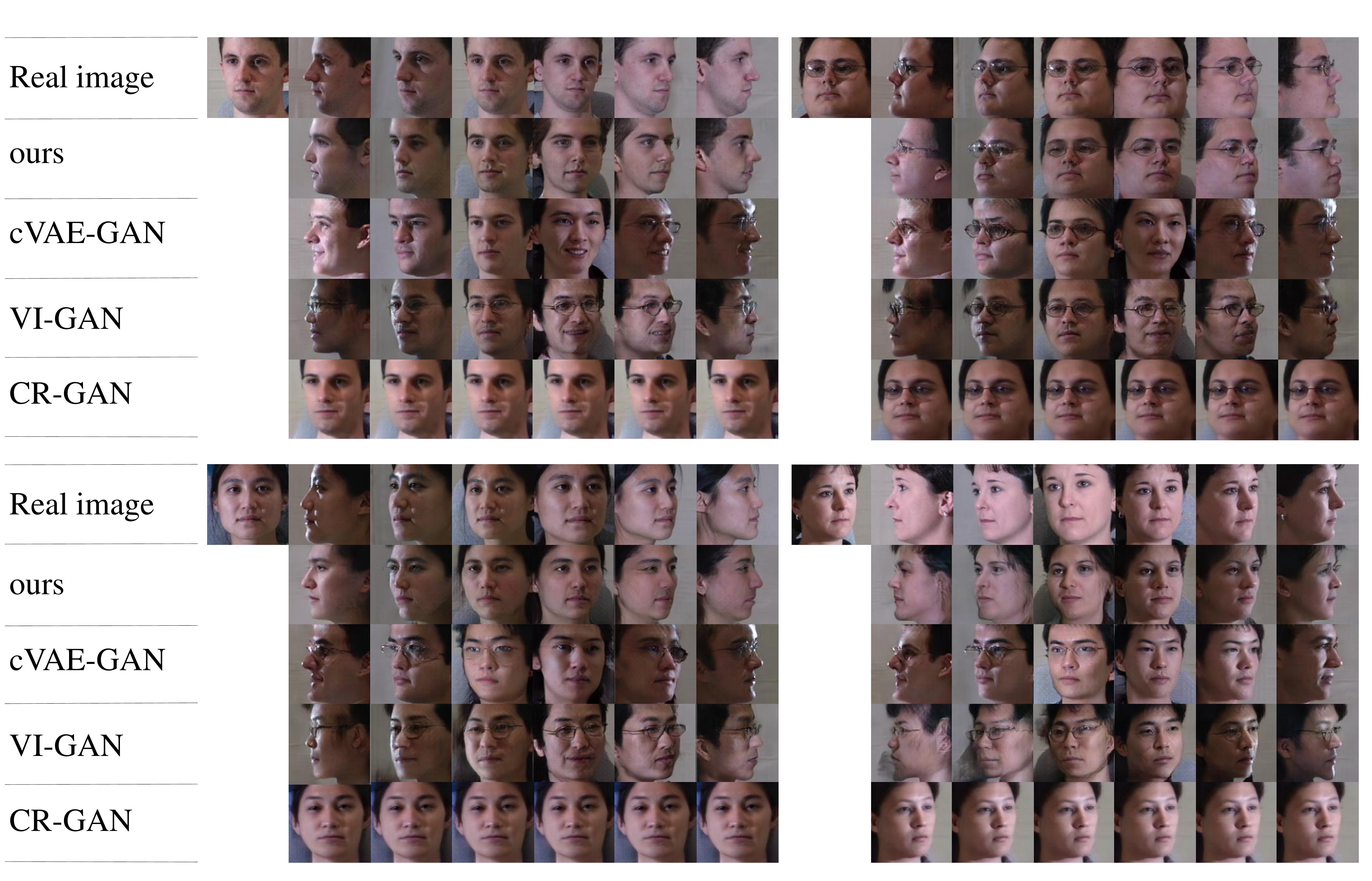}
\caption{ Comparison on Multi-PIE. For each image, the top row is the ground truth while the second row is generated by ours. The third , fourth and fifth rows are the output of cVAE-GAN\cite{bao2017cvae} ,VI-GAN\cite{xu2019view} and CR-GAN\cite{tian2018cr} respectively. }
\label{fig:comp}
\vspace{-0.5cm}
\end{figure}

\textbf{View-translation among discrete angles. }
Qualitative comparisons are performed among our proposed method and the existing works like cVAE-GAN \cite{bao2017cvae}, VI-GAN \cite{xu2019view} and CR-GAN \cite{tian2018cr}. The results are listed in Figure \ref{fig:comp}. Note that in this study, we do not use paired data for all experiments during training. 
The results of the quantitative metrics on each method are shown in the Table \ref{Table2}. After removing the constraint from the paired data, CR-GAN can hardly realize the view translation. The image qualities of VI-GAN significantly deteriorate under the condition of large angle translation. Although cVAE-GAN can still work, the converted image can not keep the view-irrelevant details from the source.
\begin{table}[] 
\begin{center}
\setlength{\tabcolsep}{1mm}{
\begin{tabular}{ l l l c c c c c c c}
\toprule
& method && \multicolumn{3}{c}{MultiPIE}&&\multicolumn{3}{c }{3D chair}  \\
\cline{4-6}\cline{8-10}
&&&L1  &SSIM &FID&  &L1  &SSIM &FID  \\
\midrule						 	
& CR-GAN\cite{tian2018cr}&             & $39.80$  & $0.397$ &$48.87$&   &$13.45$    & $0.696$ &$111.34$ \\	
&VI-GAN\cite{xu2019view}&      	  	  & $38.18$  & $0.464$   &$47.02$&   & $10.54$  & $0.802$ &$105.78$  \\		
&cVAE-GAN\cite{bao2017cvae}&     	  & $31.37$  & $0.493$ &$44.84$&    & $8.39$   & $0.859$ &$104.78$  \\			
& Ours&     & \boldmath{ $21.90$ }  & \boldmath{$0.571$} &\boldmath{$23.95$}& & \boldmath{$6.37$} &\boldmath{ $0.885$} &\boldmath{$86.34$}   	\\
\bottomrule
\end{tabular}
}
\end{center}
\caption{Quantitative metrics comparisons. Results from CR-GAN, VI-GAN and cVAE-GAN are provided on MultiPIE and the 3D chair datasets, respectively.}
\vspace{-1cm}
\label{Table2}
\end{table}

\textbf{Continuous view synthesis by interpolation. 
} 
Synthesizing images at continuously varying angles is important in real applications. In our implementation, this can be achieved by interpolating between two adjacent labels. Meanwhile, we realize that the filter $W$, computed from the discrete view labels through the MLPs $\Psi$, can help synthesizing the image at an unseen angle. Therefore, we can also directly interpolate on $W$. 
 \begin{figure}
\centering
\includegraphics[width=1.0\textwidth]{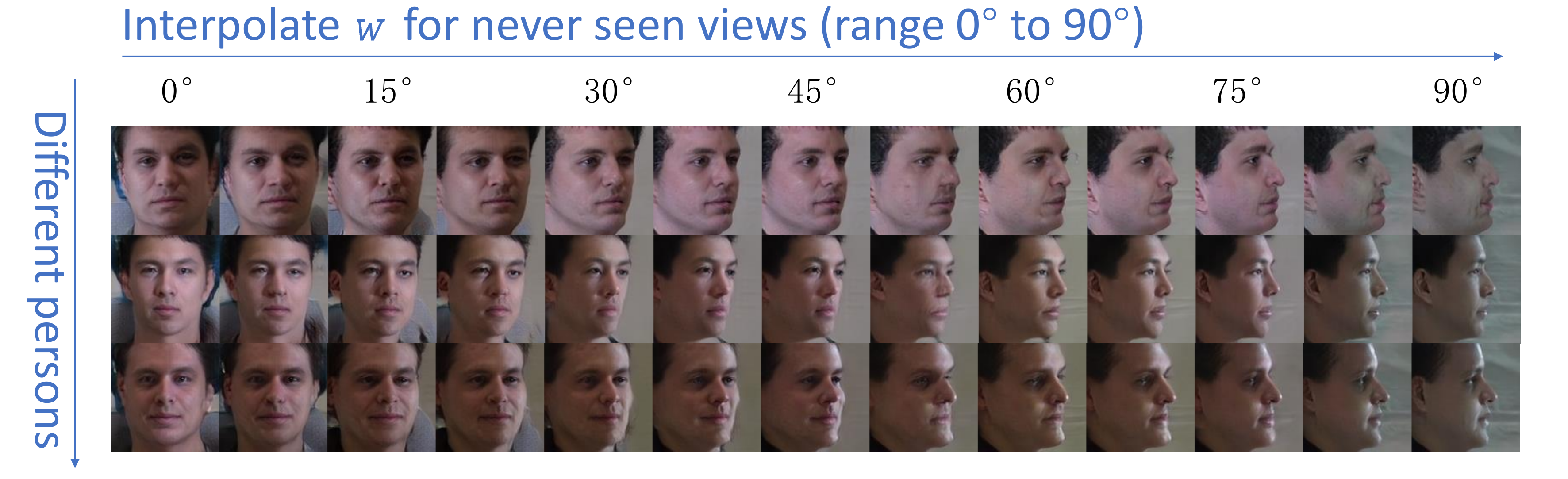}
\caption{Interpolating $W$ to synthesis unseen view images.}
\label{fig:interpolation}
\vspace{-0.5cm}
\end{figure}
The minimum angle interval in MultiPIE is $15^\circ$, and we choose to interpolate at every $7.5^\circ$. As is shown in Figure \ref{fig:interpolation}, we visualize all the images by interpolating $W$ from $0^\circ$ to 90$\circ$ and find that the face realized smooth transformation. 

\begin{figure}
\vspace{-1cm}
\centering
\includegraphics[width=0.8\textwidth]{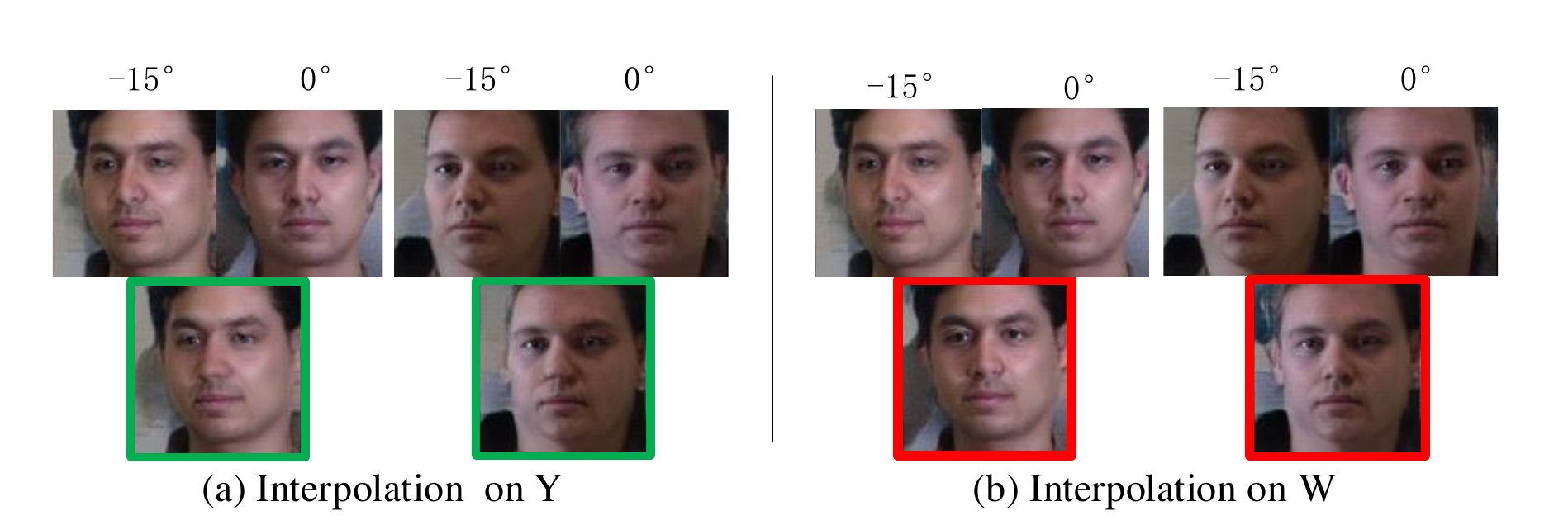}
\caption{Comparisons on different interpolation schemes for synthesizing an unseen view image on MultiPIE.}
\label{fig:interpolation sub}
\vspace{-1cm}
\end{figure}

For comparison, zooming-in results by interpolating on both $W$ and $Y$ are given in Figure \ref{fig:interpolation sub}.  
Note that all these images are the outputs from our model with the source view at $0^\circ$. The image marked with the red box is the obtained by interpolating $W$, while the green box is the result from interpolating $Y$. The results show that interpolation on $W$ gives the more accurate images. This also demonstrates that we have learned good representation $W$ for the angle since it directly relates to the optical flow on the feature.
The above results can be verified by the quantitative metric of FID. By interpolation on $W$, FID achieves $30.70$, while it is $32.04$ if the interpolation is implemented on $Y$. 

\section{Conclusions}\label{sec:con}
This paper proposes the conditional deformable VAE for the novel view synthesis based on unpaired training data. We design the CDM and DFNM which are utilized in both the encoder and decoder. The CDM employs the latent code mapping from the conditional view label as the filters to convolve the feature, so that a set of optical flows can be obtained to deform the features. The output from CDM are not directly given to the later layers, instead, they take effect through DFNM, which actually performs the conditional normalization according to its input. The experiments on 3D chair and MultiPIE show the effectiveness of our method particularly for unpaired training. 

\appendix
\section{Appendices}
In this supplementary, we first give the detailed network structures and then provide more supported results.
\subsection{Details about network structures}
\subsubsection{The structure of $G$ on 3D chair.}
\begin{figure}
\centering
\includegraphics[width=1.0\textwidth]{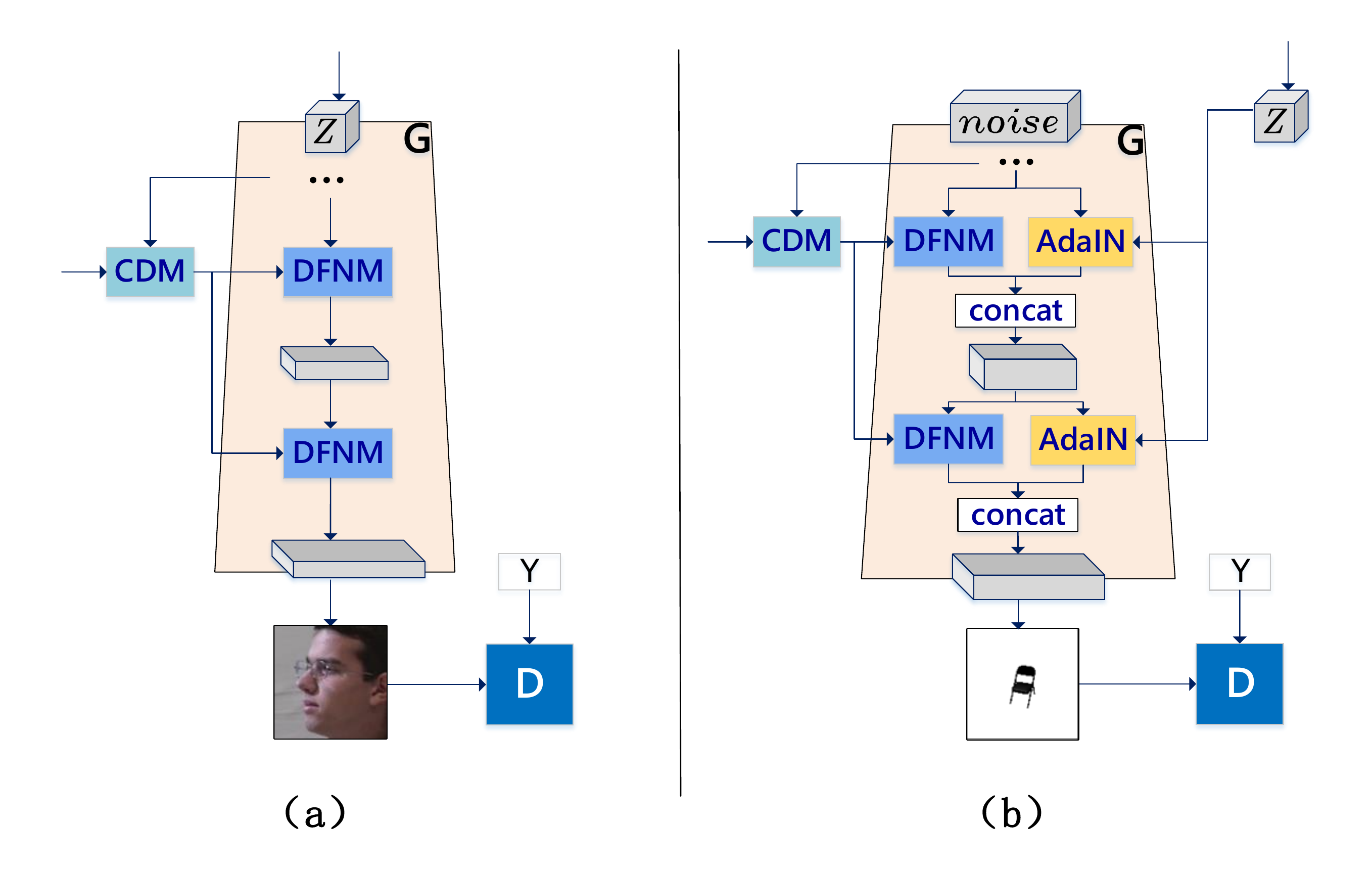}
\caption{The two different proposed structures for $G$. (a) The structure on MultiPIE. (b) The structure on 3D chair.}
\label{fig:main1 chair}
\end{figure}

The decoder structure of $G$ on the 3D chair is shown in Figure \ref{fig:main1 chair} (b). Different from the structure for MultiPIE shown in Figure \ref{fig:main1 chair} (a), the $Z$ sampled from the posterior $E(Z|X_a, Y_a)$ or prior $N(0, I)$ is given to the network from the side branch by AdaIN. The features in the main branch are affected by two outputs from the side branches through the AdaIN and DFNM, respectively. 
Their results are then concatenated along the channel dimension, and given to the next layer. 

\subsubsection{Detailed structures of different modules. }
Table \ref{tab:encoder}, \ref{tab:generator}, \ref{tab:discriminator}, \ref{tab:cdm}, \ref{tab:mappinp}, \ref{tab:dac} are the specific structures of the network of $E$, $G$, $D$, CDM, $\Psi$, and DAC, respectively,  $\to$  means directly given. Note that the decoder $G$ in Table \ref{tab:generator} is used on MuliPIE. 

\begin{table*}[htb]
\begin{center}
\setlength{\tabcolsep}{2mm}{
\begin{tabular}{|c|}
\hline
Encoder $E$\\
\hline\hline
 input ${X} \in \mathbb{R}^{128 \times 128 \times 3}$\\
 \hline
conv (F:32, K:7, S:1), IN, lrelu\\
 \hline
conv (F:64, K:4, S:2), IN, lrelu  \\
 \hline
conv (F:128, K:4, S:2), IN, lrelu \\
 \hline
conv (F:128, K:4, S:2), IN, lrelu\\
 \hline
input $Y \to$ $\Psi_E \to$ CDM\\
 \hline
 Residual block: conv (F:128, K:3, S:1), DFNM, lrelu \\
 \hline
 Residual block: conv (F:128, K:3, S:1), DFNM, lrelu \\ 
 \hline 
 Residual block: conv (F:128, K:4, S:1), DFNM, lrelu \\
 \hline
conv (F:128, K:4, S:2), IN, lrelu\\
 \hline
conv (F:128, K:3, S:2), IN, lrelu\\
 \hline
fc, 1024, lrelu\\
 \hline
 fc, 256 ($\mu$), fc, 256 ($\Sigma$)\\
 \hline

\end{tabular}
}
\end{center}
\caption{The structure of the encoder $E$. Note that the proposed CDM is used at the beginning of the first residual block to inject the input condition into the main branch. Details about CDM and $\Psi$ are given in the following tables.}
\label{tab:encoder}
\end{table*}

\makeatletter\def\@captype{table}\makeatother
\begin{center}
\setlength{\tabcolsep}{2mm}{
\begin{tabular}{|c|c|}
\hline
 Decoder $G$\\
\hline\hline
 input ${Z\in \mathbb{R}^{256}}$\\
 \hline
conv (F:128, K:3, S:1), lrelu\\
 \hline
conv (F:128, K:4, S:2), lrelu  \\
 \hline
conv (F:128, K:4, S:2), lrelu \\
 \hline
input $Y'\to \Psi_G \to$ CDM\\
 \hline
Residual block: conv (F:128, K:3, S:1), DFNM, lrelu  \\
 \hline
Residual block: conv (F:128, K:3, S:1), DFNM, lrelu\\
 \hline
Residual block: conv (F:128, K:3, S:1), DFNM, lrelu\\ 
 \hline 
conv (F:128, K:4, S:1), LN, lrelu\\
 \hline
conv (F:128, K:4, S:1), LN, lrelu\\
 \hline
conv (F:64, K:4, S:1), LN, lrelu\\
 \hline
conv (F:32, K:7, S:1), LN, lrelu  \\
 \hline
conv (F:3, K:1, S:1), tanh\\
 \hline

\end{tabular}
}
\end{center}
\caption{The structure of decoder $G$ for MultiPIE dataset.}
\label{tab:generator}

\makeatletter\def\@captype{table}\makeatother
\begin{center}
\setlength{\tabcolsep}{2mm}{
\begin{tabular}{|c|c|}
\hline
\multicolumn{2}{|c|}{Discriminator}  \\
\hline\hline
\multicolumn{2}{|c|}{input ${ X} \in \mathbb{R}^{128 \times 128 \times 3}$} \\
\hline
\multicolumn{2}{|c|}{conv(F:64, K:1, S:1), lrelu}\\
 \hline
\multicolumn{2}{|c|}{Residual block: conv (F:128, K:3, S:1), SN, lrelu} \\ 
 \hline 
\multicolumn{2}{|c|}{ downsample}  \\
  \hline
\multicolumn{2}{|c|}{Residual block: conv (F:128, K:3, S:1), SN, lrelu} \\ 
 \hline 
\multicolumn{2}{|c|}{ downsample} \\
  \hline
\multicolumn{2}{|c|}{Residual block: conv (F:256, K:3, S:1), SN, lrelu} \\ 
 \hline 
 \multicolumn{2}{|c|}{downsample} \\
  \hline
 \multicolumn{2}{|c|}{Minibatch state concat\cite{karras2017progressive} }\\
  \hline
 \multicolumn{2}{|c|}{conv (F:256, K:3, S:1), SN, lrelu} \\
 \hline
\multicolumn{2}{|c|}{ conv (F:256, K:4, S:1), SN, lrelu} \\
 \hline
 input $Y \to $fc, SN $\to$ Inner product &  fc (1), SN \\ 
 \hline
 \multicolumn{2}{|c|}{add}\\
 \hline 
\end{tabular}
}
\end{center}
\caption{The structure of projection discriminator $D$. Note that "SN" indicates the spectral normalization. }
\label{tab:discriminator}

\makeatletter\def\@captype{table}\makeatother
\begin{center}
\begin{tabular}{|c|}
\hline
 CDM \\
\hline\hline
 input $  F  \in \mathbb{R}^{16 \times 16 \times 128}$ \\ 
\hline
 conv (F:9$\times$25, K:3, S:1)  \\
\hline
input $W_y \in \mathbb{R}^{1\times1\times25} \to $KGconv (F:9, K:1, S:1)$\to dy$\\  
input $W_x \in \mathbb{R}^{1\times1\times25} \to $KGconv (F:9, K:1, S:1)$\to dx$\\
concat\\
\hline
 deformable conv (F:128, K:3, S:1) input $F$  \\
\hline
 concat input $F$ \\
\hline
\end{tabular}
\end{center}
\caption{The network structure of CDM.}
\label{tab:cdm}

\makeatletter\def\@captype{table}\makeatother
\begin{center}
\begin{tabular}{|c|}
\hline
 $\Psi$ \\
\hline\hline
 input ${\bf Y} $\\
 \hline
 fc(128) \\
 \hline
 concat ${\bf noise} \in \mathbb{R}^{128}$ \\
 \hline
 pixel norm  \\
 \hline
 fc(256) \\
\hline
\end{tabular}
\end{center}
\caption{The network structure of $\Psi$.}
\label{tab:mappinp}

\makeatletter\def\@captype{table}\makeatother
\begin{center}
\begin{tabular}{|c|}
\hline
DAC \\
\hline\hline
input ${\bf Z} \in \mathbb{R}^{256}$\\
\hline
fc(256) \\
\hline
fc(13 on MultiPIE) or (62 on 3D chair)  \\
\hline
\end{tabular}
\end{center}
\caption{The network structure of DAC.}
\label{tab:dac}

\subsection{Additional results and analysis}
\subsubsection{View synthesis from the paired data training.}
In this section, we provide the synthesis results obtained from the model trained by the paired data, which means that the target view image is directly used to constrain the model.
\begin{figure}[H]
\centering
\includegraphics[width=0.7\textwidth]{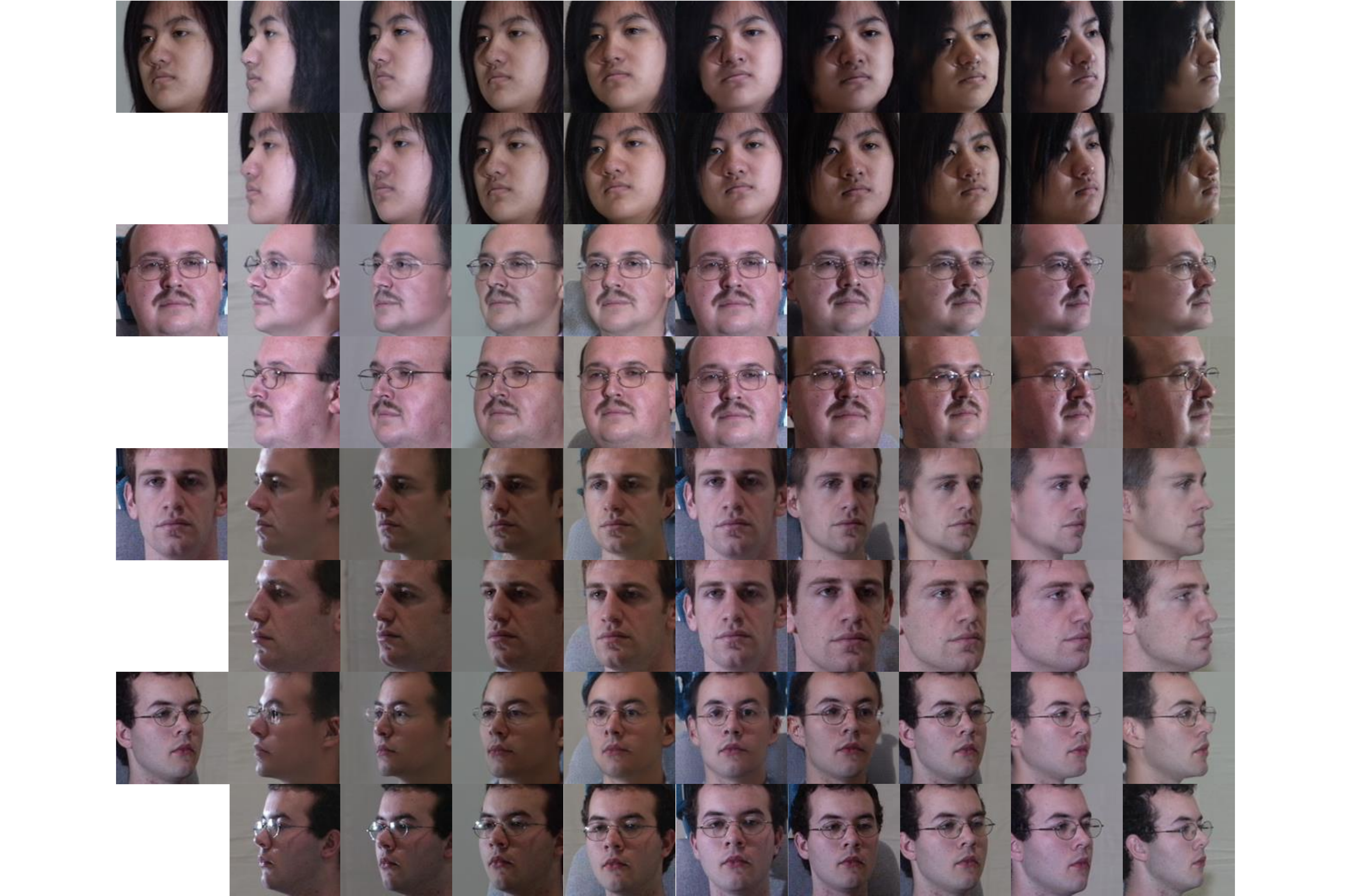}  
\caption{Synthesis results on MultiPIE training by the paired data. The first row shows the view-translated images, and the second row are the target images.  
}
\label{fig: Visualizing optical flow}
\end{figure}
\begin{table}[H] 
\begin{center}
\setlength{\tabcolsep}{1mm}{
\begin{tabular}{ l l l c c c c }
\toprule
& method && \multicolumn{3}{c}{MultiPIE}&  \\
\cline{4-6} 
&&&L1  &SSIM &FID&  \\
\midrule						 	
& our-paired& & $13.34$  & $0.63$ &$23.52$& \\ 
\bottomrule
\end{tabular}
}
\end{center}
\caption{Quantitative result on the MultiPIE based on the paired training data.}
\label{Table1}
\end{table}


\subsubsection{Visualization and analysis of the conditional flows.}
As is shown in the Figure \ref{fig: Visualizing optical flow}, we visualize the $dx$ and $dy$ used for deformation. The first row are the input image and the synthesis images under 13 different azimuths. The second row are the optical flows used by the CDM module in $E$. The third, fourth and fifth rows are $dy$, $dx$ and their differences, respectively. The sixth, seventh, eighth and ninth are the optical flow, $dy$, $dx$ and their differences in the CDM of $G$. 

Since the flows are actually the  intermediate features, they are hard to interpret. But it can be seen that when the input $Y$ of the encoder $E$ and decoder $G$ are the same, that is, the source and target view labels 
are the same, the two flows are opposite. Note that the flows in $E$ are similar since they are determined mainly by the source input $X$ and its label $Y$. The slightly differences are caused by the sampling noises introduced in $\Psi$. 
\begin{figure}[H]
\centering
\includegraphics[width=1.0\textwidth]{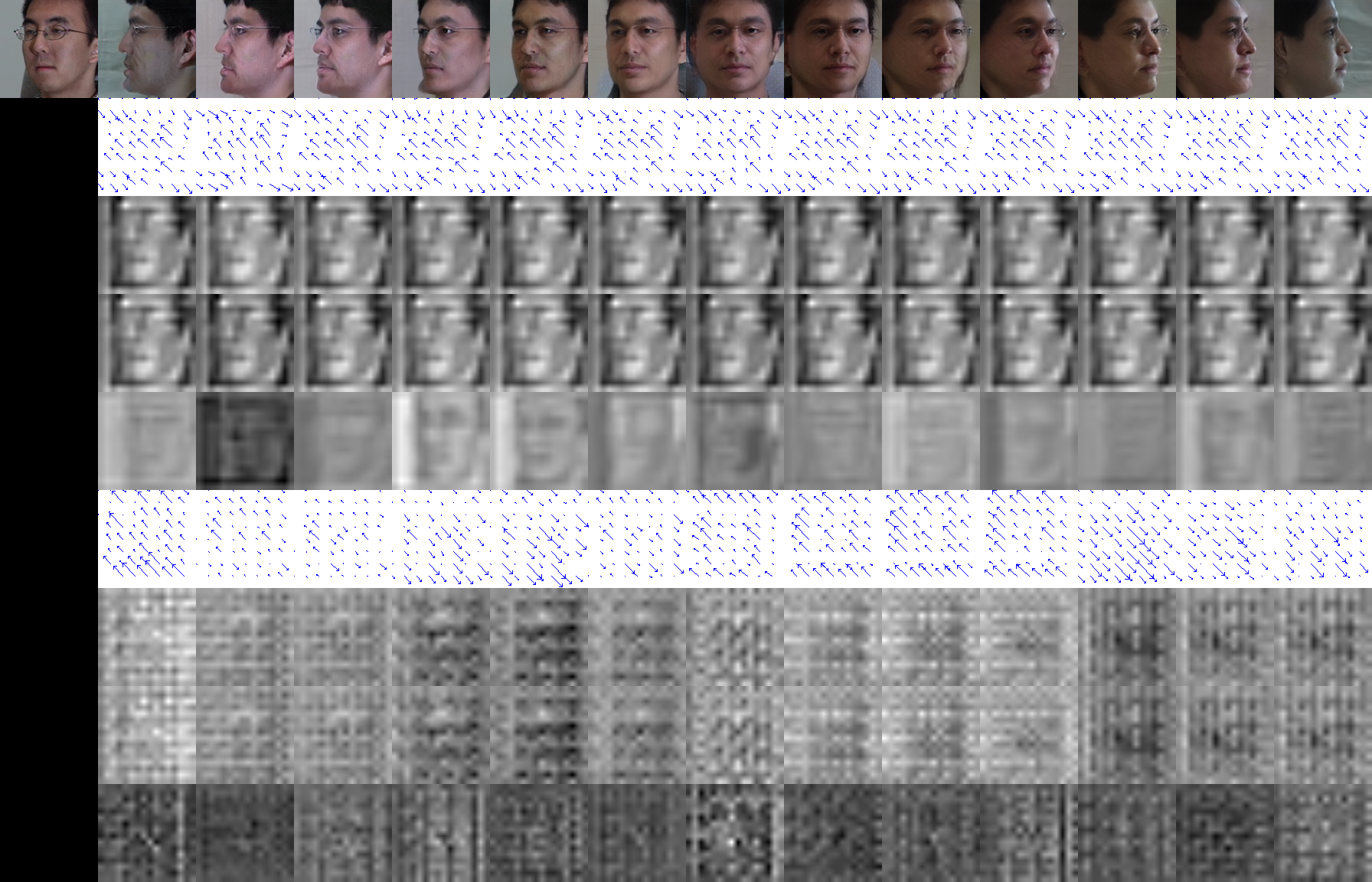} 
\includegraphics[width=1.0\textwidth]{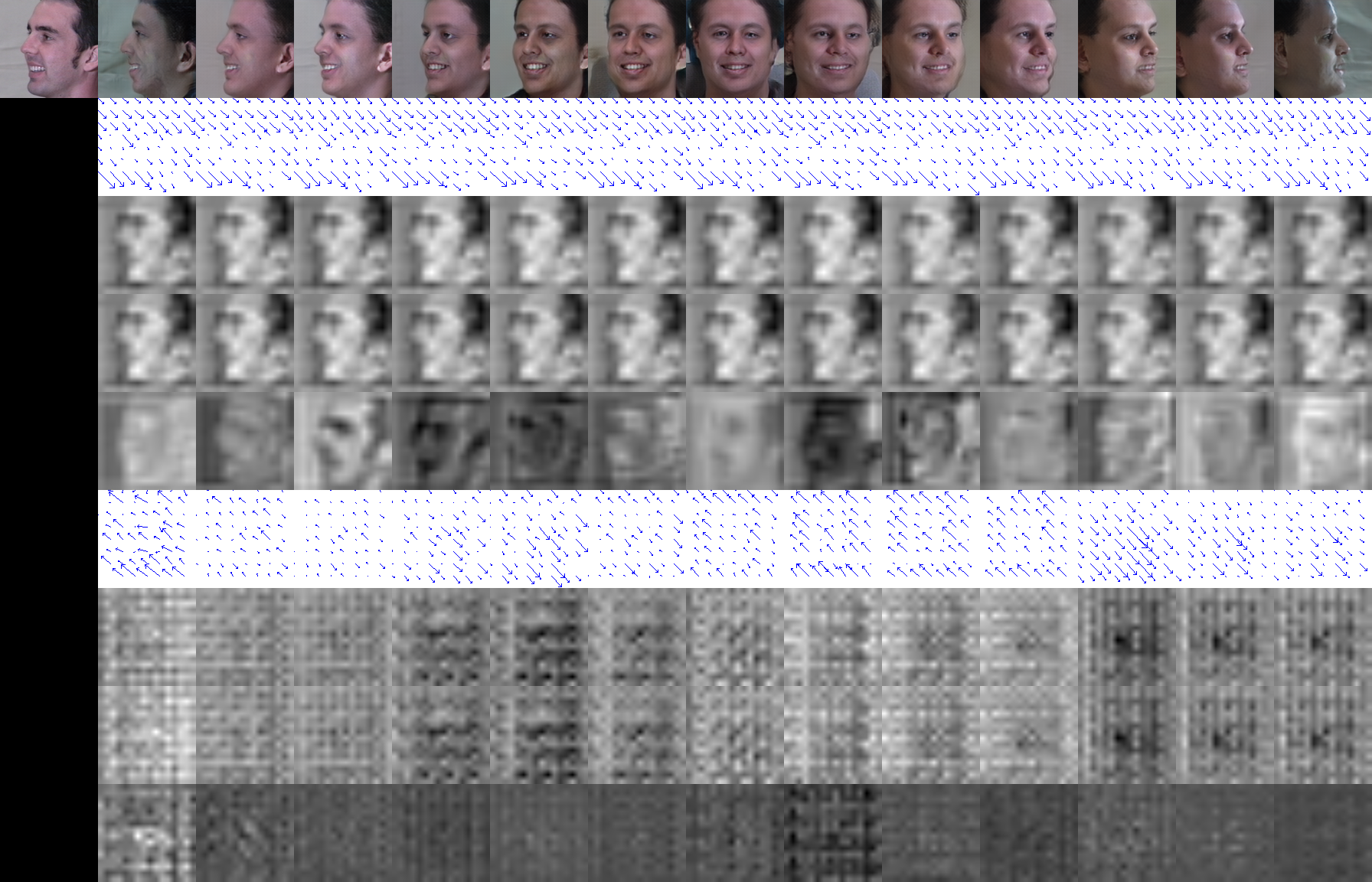}  
\caption*{}
\label{fig: Visualizing optical flow}
\end{figure}

\begin{figure}[H]
\centering  
\includegraphics[width=1.0\textwidth]{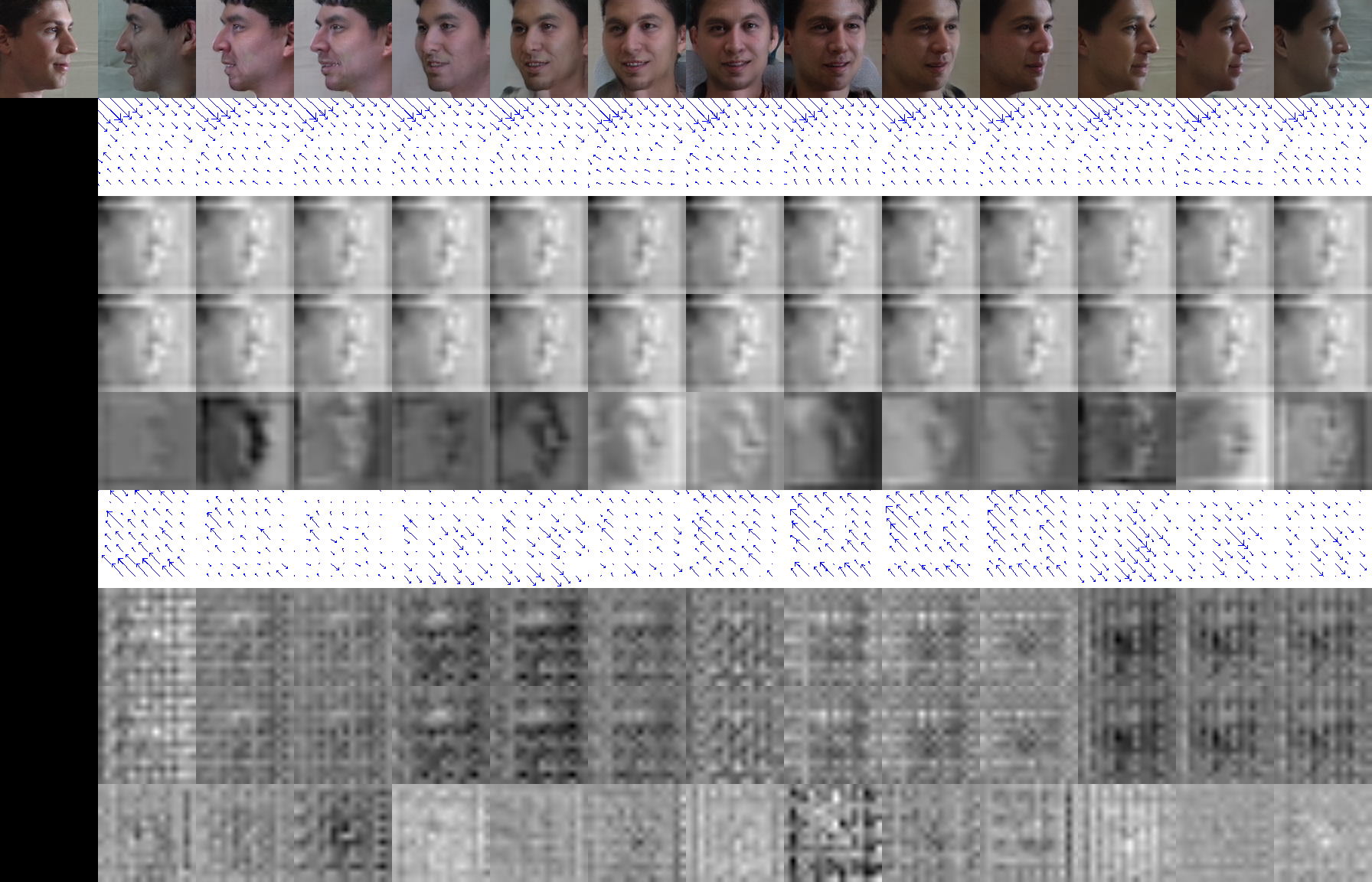} 
\caption{ Visualization of the optical flow and its the two components $dx$ and $dy$ under different source and target view combinations. The first row shows the input source image and its different target view translation results. The second and sixth rows are the optical flows in the CDM of $E$ and $G$. The third and fourth rows are $dy$ and $dx$ components in $E$, while fifth row explicitly shows their differences. $dy$, $dx$ and their differences in $G$ are shown in the seventh, eighth and ninth rows.
}
\label{fig: Visualizing optical flow}
\end{figure}

\subsubsection{Additional experiment results} 
\begin{figure}[H]
\centering
\includegraphics[width=1.0\textwidth]{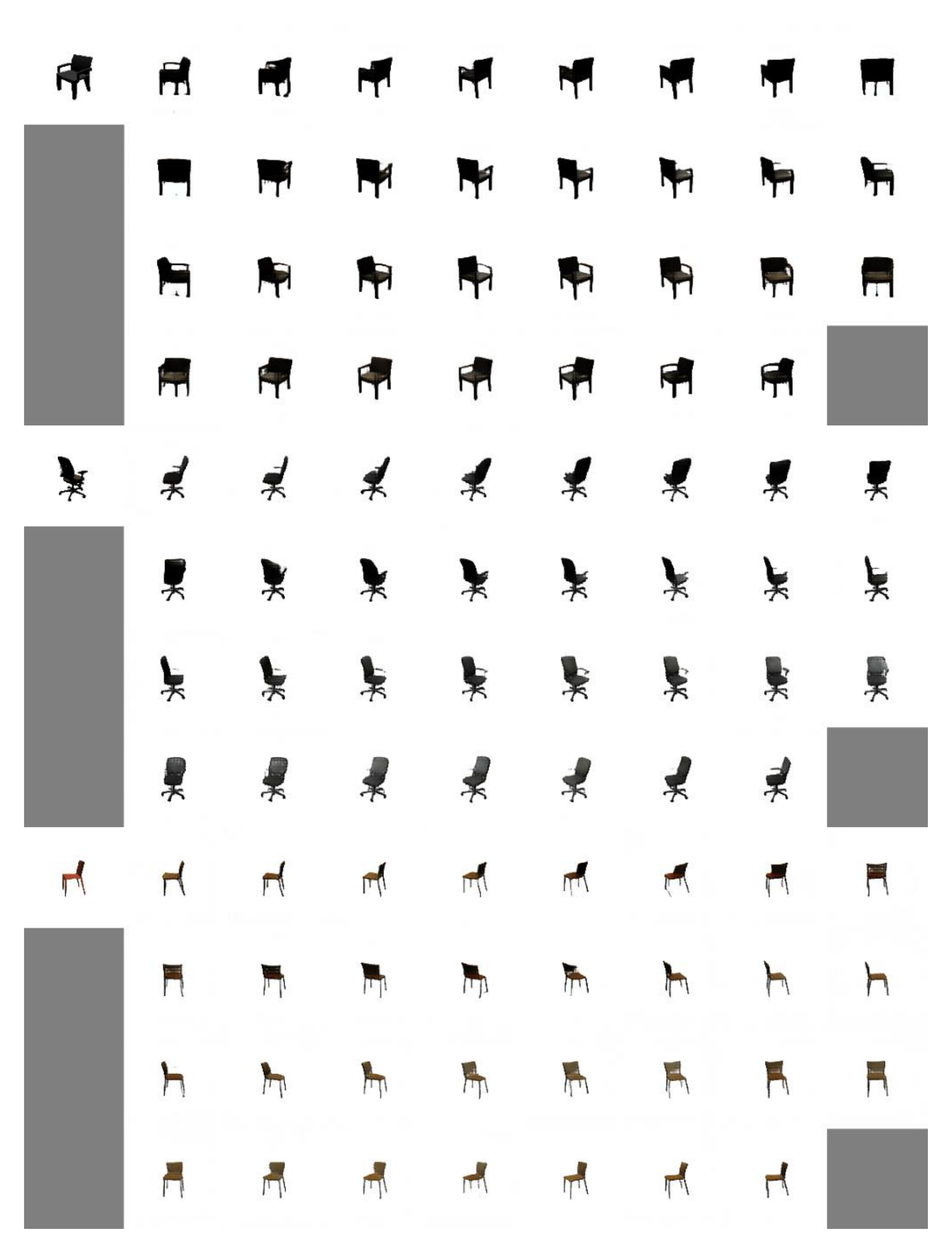}
\caption{Synthesized images of different views on 3D chair dataset.The first one is input image, and the remaining are generated images under 31 different views.}
\label{fig:sup_chair_allview1}
\end{figure}

\begin{figure}[H]
\centering
\includegraphics[width=1.0\textwidth]{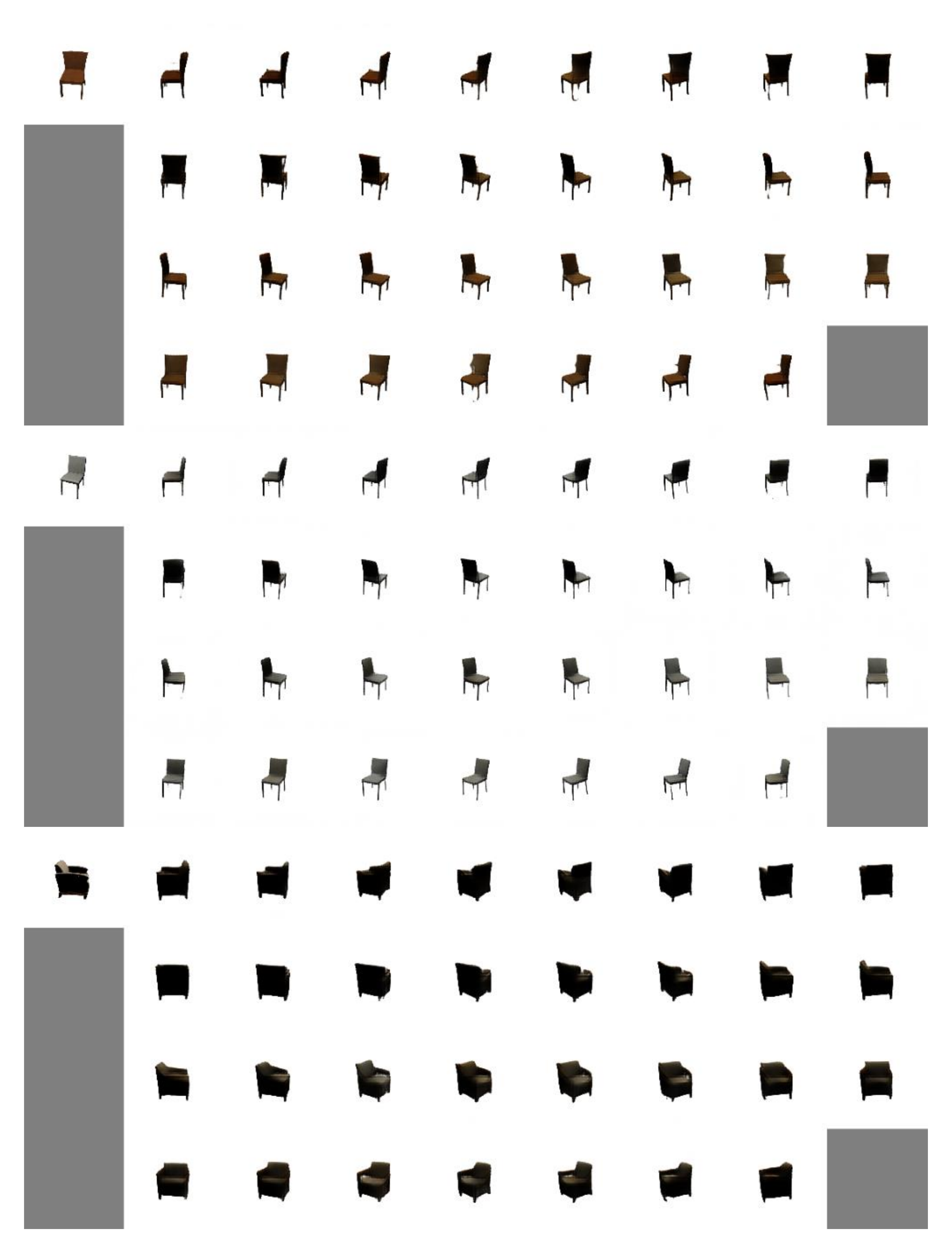} 
\caption{Synthesized images of different views on 3D chair dataset.The first one is input image, and the remaining are generated images under 31 different views.}
\label{fig:sup_chair_allview2}
\end{figure}

\begin{figure}[H]
\centering
\includegraphics[width=1.0\textwidth]{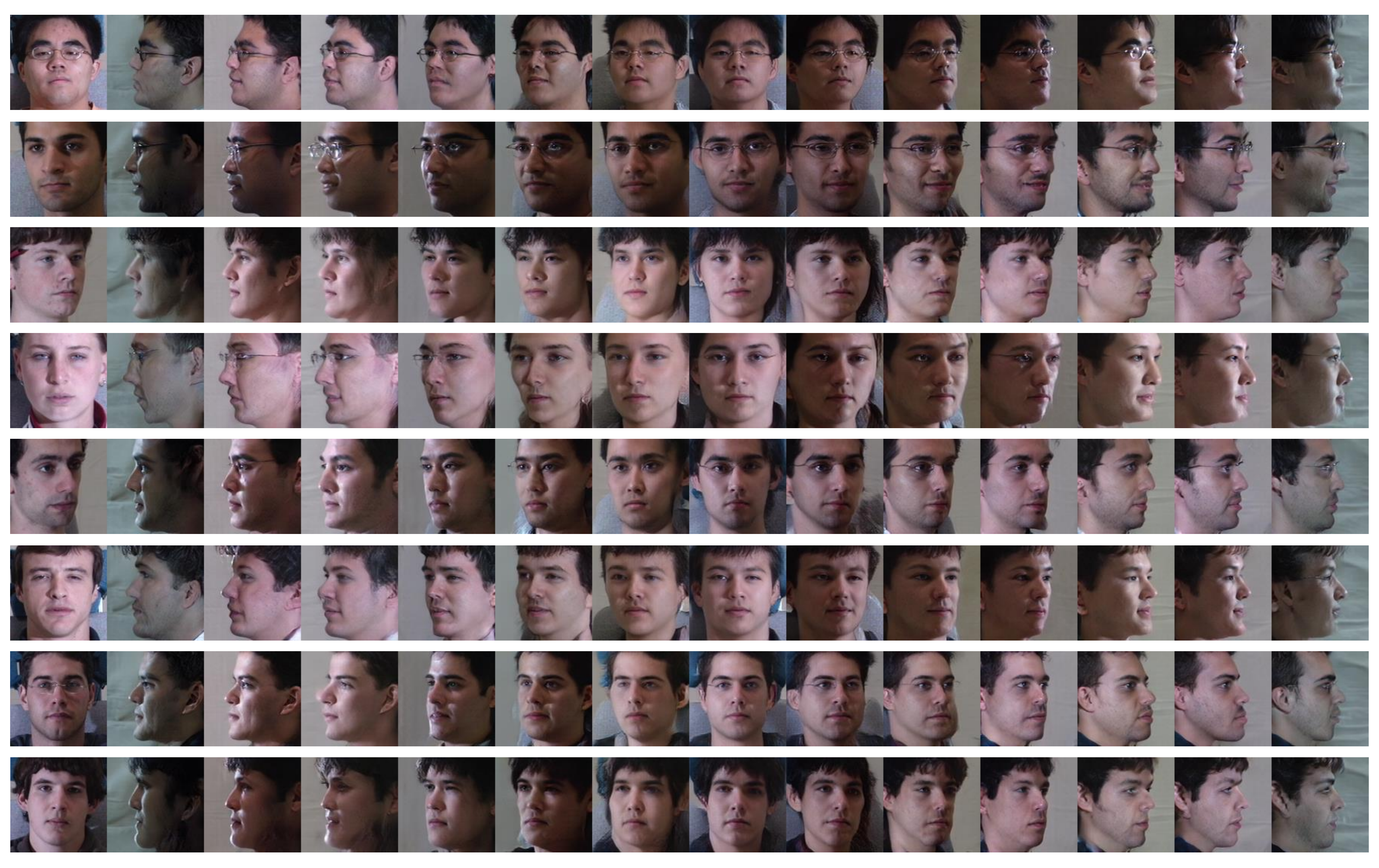} 
\caption{Synthesized images of different views on MultiPIE dataset. The first column is input image, and the remaining 13 columns are view-translated images under 13 different target views.}
\label{fig:Ablation_studies_fasion}
\end{figure}

on two datasets are provided in Figure \ref{fig:sup_chair_allview1}, \ref{fig:sup_chair_allview2} and \ref{fig:Ablation_studies_fasion}.


%
%
%
%
\bibliographystyle{abbrv}
\bibliography{egbib}

\begin{thebibliography}{10}

\bibitem{aubry2014seeing}
M.~Aubry, D.~Maturana, A.~A. Efros, B.~C. Russell, and J.~Sivic.
\newblock Seeing 3d chairs: exemplar part-based 2d-3d alignment using a large
  dataset of cad models.
\newblock In {\em Proceedings of the IEEE conference on computer vision and
  pattern recognition}, pages 3762--3769, 2014.

\bibitem{avidan1997novel}
S.~Avidan and A.~Shashua.
\newblock Novel view synthesis in tensor space.
\newblock In {\em Proceedings of IEEE Computer Society Conference on Computer
  Vision and Pattern Recognition}, pages 1034--1040. IEEE, 1997.

\bibitem{bao2017cvae}
J.~Bao, D.~Chen, F.~Wen, H.~Li, and G.~Hua.
\newblock Cvae-gan: fine-grained image generation through asymmetric training.
\newblock In {\em Proceedings of the IEEE International Conference on Computer
  Vision}, pages 2745--2754, 2017.

\bibitem{dai2017deformable}
J.~Dai, H.~Qi, Y.~Xiong, Y.~Li, G.~Zhang, H.~Hu, and Y.~Wei.
\newblock Deformable convolutional networks.
\newblock In {\em Proceedings of the IEEE international conference on computer
  vision}, pages 764--773, 2017.

\bibitem{dosovitskiy2015learning}
A.~Dosovitskiy, J.~Tobias~Springenberg, and T.~Brox.
\newblock Learning to generate chairs with convolutional neural networks.
\newblock In {\em Proceedings of the IEEE Conference on Computer Vision and
  Pattern Recognition}, pages 1538--1546, 2015.

\bibitem{goodfellow2014generative}
I.~Goodfellow, J.~Pouget-Abadie, M.~Mirza, B.~Xu, D.~Warde-Farley, S.~Ozair,
  A.~Courville, and Y.~Bengio.
\newblock Generative adversarial nets.
\newblock In {\em Advances in neural information processing systems}, pages
  2672--2680, 2014.

\bibitem{gross2010multi}
R.~Gross, I.~Matthews, J.~Cohn, T.~Kanade, and S.~Baker.
\newblock Multi-pie.
\newblock {\em Image and Vision Computing}, 28(5):807--813, 2010.

\bibitem{gulrajani2017improved}
I.~Gulrajani, F.~Ahmed, M.~Arjovsky, V.~Dumoulin, and A.~C. Courville.
\newblock Improved training of wasserstein gans.
\newblock In {\em Advances in neural information processing systems}, pages
  5767--5777, 2017.

\bibitem{heusel2017gans}
M.~Heusel, H.~Ramsauer, T.~Unterthiner, B.~Nessler, and S.~Hochreiter.
\newblock Gans trained by a two time-scale update rule converge to a local nash
  equilibrium.
\newblock In {\em Advances in neural information processing systems}, pages
  6626--6637, 2017.

\bibitem{higgins2017beta}
I.~Higgins, L.~Matthey, A.~Pal, C.~Burgess, X.~Glorot, M.~Botvinick,
  S.~Mohamed, and A.~Lerchner.
\newblock beta-vae: Learning basic visual concepts with a constrained
  variational framework.
\newblock {\em Iclr}, 2(5):6, 2017.

\bibitem{huang2017arbitrary}
X.~Huang and S.~Belongie.
\newblock Arbitrary style transfer in real-time with adaptive instance
  normalization.
\newblock In {\em Proceedings of the IEEE International Conference on Computer
  Vision}, pages 1501--1510, 2017.

\bibitem{isola2017image}
P.~Isola, J.-Y. Zhu, T.~Zhou, and A.~A. Efros.
\newblock Image-to-image translation with conditional adversarial networks.
\newblock In {\em Proceedings of the IEEE conference on computer vision and
  pattern recognition}, pages 1125--1134, 2017.

\bibitem{kholgade20143d}
N.~Kholgade, T.~Simon, A.~Efros, and Y.~Sheikh.
\newblock 3d object manipulation in a single photograph using stock 3d models.
\newblock {\em ACM Transactions on Graphics (TOG)}, 33(4):1--12, 2014.

\bibitem{kingma2014adam}
D.~P. Kingma and J.~Ba.
\newblock Adam: A method for stochastic optimization.
\newblock {\em arXiv preprint arXiv:1412.6980}, 2014.

\bibitem{kingma2013auto}
D.~P. Kingma and M.~Welling.
\newblock Auto-encoding variational bayes.
\newblock {\em arXiv preprint arXiv:1312.6114}, 2013.

\bibitem{larsen2015autoencoding}
A.~B.~L. Larsen, S.~K. S{\o}nderby, H.~Larochelle, and O.~Winther.
\newblock Autoencoding beyond pixels using a learned similarity metric.
\newblock {\em arXiv preprint arXiv:1512.09300}, 2015.

\bibitem{massa2016deep}
F.~Massa, B.~C. Russell, and M.~Aubry.
\newblock Deep exemplar 2d-3d detection by adapting from real to rendered
  views.
\newblock In {\em Proceedings of the IEEE Conference on Computer Vision and
  Pattern Recognition}, pages 6024--6033, 2016.

\bibitem{mirza2014conditional}
M.~Mirza and S.~Osindero.
\newblock Conditional generative adversarial nets.
\newblock {\em arXiv preprint arXiv:1411.1784}, 2014.

\bibitem{miyato2018spectral}
T.~Miyato, T.~Kataoka, M.~Koyama, and Y.~Yoshida.
\newblock Spectral normalization for generative adversarial networks.
\newblock {\em arXiv preprint arXiv:1802.05957}, 2018.

\bibitem{miyato2018cgans}
T.~Miyato and M.~Koyama.
\newblock cgans with projection discriminator.
\newblock {\em arXiv preprint arXiv:1802.05637}, 2018.

\bibitem{nguyen2019hologan}
T.~Nguyen-Phuoc, C.~Li, L.~Theis, C.~Richardt, and Y.-L. Yang.
\newblock Hologan: Unsupervised learning of 3d representations from natural
  images.
\newblock In {\em Proceedings of the IEEE International Conference on Computer
  Vision}, pages 7588--7597, 2019.

\bibitem{park2017transformation}
E.~Park, J.~Yang, E.~Yumer, D.~Ceylan, and A.~C. Berg.
\newblock Transformation-grounded image generation network for novel 3d view
  synthesis.
\newblock In {\em Proceedings of the ieee conference on computer vision and
  pattern recognition}, pages 3500--3509, 2017.

\bibitem{park2019semantic}
T.~Park, M.-Y. Liu, T.-C. Wang, and J.-Y. Zhu.
\newblock Semantic image synthesis with spatially-adaptive normalization.
\newblock In {\em Proceedings of the IEEE Conference on Computer Vision and
  Pattern Recognition}, pages 2337--2346, 2019.

\bibitem{rematas2016novel}
K.~Rematas, C.~H. Nguyen, T.~Ritschel, M.~Fritz, and T.~Tuytelaars.
\newblock Novel views of objects from a single image.
\newblock {\em IEEE transactions on pattern analysis and machine intelligence},
  39(8):1576--1590, 2016.

\bibitem{rematas2014image}
K.~Rematas, T.~Ritschel, M.~Fritz, and T.~Tuytelaars.
\newblock Image-based synthesis and re-synthesis of viewpoints guided by 3d
  models.
\newblock In {\em Proceedings of the IEEE Conference on Computer Vision and
  Pattern Recognition}, pages 3898--3905, 2014.

\bibitem{savarese2008view}
S.~Savarese and L.~Fei-Fei.
\newblock View synthesis for recognizing unseen poses of object classes.
\newblock In {\em European Conference on Computer Vision}, pages 602--615.
  Springer, 2008.

\bibitem{shepard1971mental}
R.~N. Shepard and J.~Metzler.
\newblock Mental rotation of three-dimensional objects.
\newblock {\em Science}, 171(3972):701--703, 1971.

\bibitem{sohn2015learning}
K.~Sohn, H.~Lee, and X.~Yan.
\newblock Learning structured output representation using deep conditional
  generative models.
\newblock In {\em Advances in neural information processing systems}, pages
  3483--3491, 2015.

\bibitem{sun2018multi}
S.-H. Sun, M.~Huh, Y.-H. Liao, N.~Zhang, and J.~J. Lim.
\newblock Multi-view to novel view: Synthesizing novel views with self-learned
  confidence.
\newblock In {\em Proceedings of the European Conference on Computer Vision
  (ECCV)}, pages 155--171, 2018.

\bibitem{tian2018cr}
Y.~Tian, X.~Peng, L.~Zhao, S.~Zhang, and D.~N. Metaxas.
\newblock Cr-gan: learning complete representations for multi-view generation.
\newblock {\em arXiv preprint arXiv:1806.11191}, 2018.

\bibitem{tran2017disentangled}
L.~Tran, X.~Yin, and X.~Liu.
\newblock Disentangled representation learning gan for pose-invariant face
  recognition.
\newblock In {\em Proceedings of the IEEE conference on computer vision and
  pattern recognition}, pages 1415--1424, 2017.

\bibitem{wang2014cross}
J.~Wang, X.~Nie, Y.~Xia, Y.~Wu, and S.-C. Zhu.
\newblock Cross-view action modeling, learning and recognition.
\newblock In {\em Proceedings of the IEEE Conference on Computer Vision and
  Pattern Recognition}, pages 2649--2656, 2014.

\bibitem{wang2004image}
Z.~Wang, A.~C. Bovik, H.~R. Sheikh, and E.~P. Simoncelli.
\newblock Image quality assessment: from error visibility to structural
  similarity.
\newblock {\em IEEE transactions on image processing}, 13(4):600--612, 2004.

\bibitem{xu2019view}
X.~Xu, Y.-C. Chen, and J.~Jia.
\newblock View independent generative adversarial network for novel view
  synthesis.
\newblock In {\em Proceedings of the IEEE International Conference on Computer
  Vision}, pages 7791--7800, 2019.

\bibitem{zhang2015meshstereo}
C.~Zhang, Z.~Li, Y.~Cheng, R.~Cai, H.~Chao, and Y.~Rui.
\newblock Meshstereo: A global stereo model with mesh alignment regularization
  for view interpolation.
\newblock In {\em Proceedings of the IEEE International Conference on Computer
  Vision}, pages 2057--2065, 2015.

\bibitem{zhou2016view}
T.~Zhou, S.~Tulsiani, W.~Sun, J.~Malik, and A.~A. Efros.
\newblock View synthesis by appearance flow.
\newblock In {\em European conference on computer vision}, pages 286--301.
  Springer, 2016.

\bibitem{zhu2017unpaired}
J.-Y. Zhu, T.~Park, P.~Isola, and A.~A. Efros.
\newblock Unpaired image-to-image translation using cycle-consistent
  adversarial networks.
\newblock In {\em Proceedings of the IEEE international conference on computer
  vision}, pages 2223--2232, 2017.

\bibitem{zhu2017toward}
J.-Y. Zhu, R.~Zhang, D.~Pathak, T.~Darrell, A.~A. Efros, O.~Wang, and
  E.~Shechtman.
\newblock Toward multimodal image-to-image translation.
\newblock In {\em Advances in neural information processing systems}, pages
  465--476, 2017.

\end{thebibliography}

\end{document}